%% file: main.tex
\documentclass[twocolumn, switch]{article} 

\usepackage{preprint}

\usepackage{amsmath, amsthm, amssymb, amsfonts}

\usepackage[numbers,square]{natbib}
\usepackage[utf8]{inputenc}	
\usepackage[T1]{fontenc}	
\usepackage{xcolor}		
\usepackage[colorlinks = true,
            linkcolor = purple,
            urlcolor  = blue,
            citecolor = cyan,
            anchorcolor = black]{hyperref}	
\usepackage{booktabs} 		
\usepackage{nicefrac}		
\usepackage{microtype}		
\usepackage{lineno}		
\usepackage{float}			

\usepackage{lipsum}		

\usepackage{newfloat}
\DeclareFloatingEnvironment[name={Supplementary Figure}]{suppfigure}
\usepackage{sidecap}
\sidecaptionvpos{figure}{c}

\usepackage{titlesec}
\titlespacing\section{0pt}{12pt plus 3pt minus 3pt}{1pt plus 1pt minus 1pt}
\titlespacing\subsection{0pt}{10pt plus 3pt minus 3pt}{1pt plus 1pt minus 1pt}
\titlespacing\subsubsection{0pt}{8pt plus 3pt minus 3pt}{1pt plus 1pt minus 1pt}

\usepackage{tikz,xcolor,hyperref}

\definecolor{lime}{HTML}{A6CE39}
\DeclareRobustCommand{\orcidicon}{
	\begin{tikzpicture}
	\draw[lime, fill=lime] (0,0)
	circle [radius=0.16]
	node[white] {{\fontfamily{qag}\selectfont \tiny ID}};
	\draw[white, fill=white] (-0.0625,0.095)
	circle [radius=0.007];
	\end{tikzpicture}
	\hspace{-2mm}
}
\foreach \x in {A, ..., Z}{\expandafter\xdef\csname orcid\x\endcsname{\noexpand\href{https://orcid.org/\csname orcidauthor\x\endcsname}
			{\noexpand\orcidicon}}
}

\usepackage{amssymb}
\usepackage{amsmath}
\usepackage{color}
\usepackage{tabu}
\usepackage{booktabs}

\usepackage{algorithm}
\usepackage[noend]{algorithmic}
\usepackage{algorithmic}

\title{Evaluating DTW Measures via a Synthesis Framework for Time-Series Data}

\usepackage{authblk}

\author[1\thanks{\tt{kishansingh0817@gmail.com}}]{Kishansingh Rajput\orcidA{}}
\author[1]{Duong Binh Nguyen\orcidB{}}
\author[1\thanks{\tt{gchen22@central.uh.edu}}]{Guoning Chen\orcidC{}}

\affil[1]{Department of Computer Science, University of Houston, Houston, TX, USA}

\begin{document}

\twocolumn[ 
  \begin{@twocolumnfalse} 

\maketitle

\begin{abstract}
\input{Content/Abstract}
\end{abstract}
\vspace{0.35cm}

  \end{@twocolumnfalse} 
] 



\section{Introduction}
\label{sec:intro}

\input{Content/Introduction}
\input{Content/Background}
\input{Content/SynthesisFramework}

\input{Content/Evaluation}

\input{Content/Applications}

\input{Content/Conclusion}



\normalsize
\bibliography{DTW_evl}


\end{document}

%% file: Content/Abstract.tex
Time series data originate from various applications that describe specific observations or quantities of interest over time. Their analysis often involves the comparison across different time-series data sequences, which in turn requires the alignment of these sequences. Dynamic Time Warping (DTW) is the standard approach to achieve an optimal alignment between two temporal signals. Different variations of DTW have been proposed to address various needs for signal alignment or classifications. However, a comprehensive evaluation of their performance in these time-series data processing tasks is lacking. Most DTW measures perform well on certain types of time-series data without a clear explanation of the reason. To address that, we propose a synthesis framework to model the variation between two time-series data sequences for comparison. Our synthesis framework can produce a realistic initial signal and deform it with controllable variations that mimic real-world scenarios. With this synthesis framework, we produce a large number of time-series sequence pairs with different but known variations, which are used to assess the performance of a number of well-known DTW measures for the tasks of alignment and classification. We report their performance on different variations and suggest the proper DTW measure to use based on the type of variations between two time-series. This is the first time such a guideline is presented for selecting a proper DTW measure. To validate our conclusion, we apply our findings to real-world applications, i.e., the detection of the formation top for the oil and gas industry and the pattern search in streamlines for flow visualization.

%% file: Content/Introduction.tex
Time-series data are generally observations taken at fixed time intervals. These may include observations by sensors or humans. In many applications, these temporal signals need to be aligned with one another, where one or more signals are considered as reference(s), and newly arriving signals are aligned with them. Dynamic Time Warping (DTW) is used to find an optimal alignment between two temporal signals. It warps one signal over another non-linearly by stretching or shrinking it along its time axis. There is an objective function, which is minimized by DTW while producing such warping. This warping is then used to align two signals. DTW has become a very popular measure in pattern matching for time-series data analysis and visualization \cite{chen2001discovering, shanker2007off}. It also leads to a number of variations (e.g., derivative DTW, weighted DTW, etc.) to address specific needs in different applications \cite{keogh2001derivative, jeong2011weighted, GORECKI20152305, salvador2007toward}.

Despite its wide application, there is a lack of a guideline for the selection of the proper DTW method for the needs of different applications, which requires a study on how different characteristics of the two to-be-aligned signals affect the performance of different DTW measures. To investigate how different characteristics of the signals affect the DTW alignments, two signals with known alignment and known characteristics (or difference) should be given, which is difficult to satisfy with the signals from the real-world applications. To address this challenge,  
in this paper, we introduce a new synthesis framework to generate pairs of time-series sequences with known and controllable variations (or differences). Our synthesis framework consists of two steps: (1) generate as realistic as possible time-series data and (2) produce a second time-series data with the known differences from the first one. To simplify the second step, we consider only phase shifting/scaling and random peak insertions (or removal) to variate the input time-series data to generate the second series. 

We apply these pairs of time-series data, generated with the proposed synthesis framework with different but known variations, to assess the performance of different DTW methods. To measure the quality of the alignment results with different DTW methods, a proper distance metric is needed. 
The distance measure used by most researchers in this field is the aggregate distance between the magnitude of matched points, since the ground truth matching of the two signals over the time axis is usually unknown. We refer to this measure as the aggregate distance over magnitude (ADM).
In this paper, since we use our synthesis framework to generate realistic time-series sequence pairs, the correct matching of these signals over the time axis is known. Thus, we introduce another matching evaluation measure, which we refer to as the Aggregate Distance Over Time (ADT), that aggregates differences in time values of matching points. For both ADM and ADT, the smaller their values are, the better the alignment is. With these two metrics, we report the performance of different DTW methods when applied to pairs of time-series data with known types of variations. This is the first time that such a guideline is reported.

To apply the reported guideline to the real-world data to validate its effectiveness, we develop a fitting framework that enables to deform one series (referred to as the reference) to the other (namely the target) using our synthesis framework. In particular, we adapt the simulated annealing to discover the best set of parameters that can generate a synthetic series from the reference. This set of parameters and their corresponding scaling/shifting and random peak insertion/removal help us characterize the type of variations between the two to-be-compared signals so that the conclusion from the aforementioned evaluation can be applied. We apply this strategy to the gamma ray logs used extensively in oil and gas industries to find out the depth of different surface formation transitions under the earth's crust.

In summary, the contributions of this work include (1) a user-controllable synthesis framework capable of generating realistic synthetic signals and signal pairs with known variations; (2) a comprehensive evaluation of a number of representative DTW measures when applied to the alignment of the signal pairs with known variations. We report our evaluation in the form of a guideline and apply the evaluation result to a number of real-world applications to assess its effectiveness. Our work fills the gap between the extensive research on inventing different variations of DTW and their proper selection for different needs of the real-world applications. 



%% file: Content/Background.tex
\section{Background}
\label{sec:DTWbackground}
 
\subsection{DTW and its variants}
\label{sec:dtwbasics}

Dynamic Time Warping (abbreviated as DTW)\cite{berndt1994using} is a widely used method to find warping/matching between two time-series sequences. 
The methodology for DTW is as follows. Consider two time-series sequences (or signals) $X = \{x_1, x_2, ...., x_m\}$ and $Y=\{y_1, y_2,.....,y_n\}$ of length $m$ and $n$, respectively. An $m \times n$ matrix is created, where each entry $(i,j)$ contains the distance between $x_i$ and $y_j$, such that $dist(i,j) = |x_i - y_j|$. In general, this distance is normalized to the second normal form. The path from $(0,0)$ to $(m,n)$ in this matrix with minimum aggregate distance is selected. This path is called the \textbf{warping path} \cite{berndt1994using}. This distance measure is based on ADM. In our evaluation, we will show both ADM and ADT. The running time of DTW is quadratic since it uses Dynamic Programming, but many attempts have already been made to reduce the runtime. Many variations of DTW have already been proposed that will be discussed later in this section. 
One of the limitations of DTW is in the features it considers. It only considers Y-axis values of the time-series data. Due to this, it may not be able to accurately align series which are slightly different (or shifting) on y-axis. One attempt to overcome this issue is the Derivative DTW (DDTW)\cite{keogh2001derivative}. DDTW is a variation of standard DTW where instead of taking raw series, a derivative of each point is taken into consideration. It makes DTW more accurate in certain cases by considering the direction information of the time-series data. 

Another limitation of DTW is that it cannot accurately align two series when there is high variance, i.e., the respective magnitudes of the matching points between two temporal signals are very different. When the warping path is too skewed, it is more likely to have a large error. To overcome this issue, Weighted DTW (WDTW)\cite{jeong2011weighted} was introduced. WDTW weights each distance value in the warping matrix before they are considered for the minimum distance path.
A high weight factor ensures the path to be diagonal and does not allow it to be skewed, while a low weight factor allows more flexibility. The weight factor is designed to be high around the central region of the series since it is believed that signals are more stable around the central region and low at the ends. Weighted DTW inspired the use of Weights in DDTW, hence Weighted Derivative DTW (WDDTW) was also proposed in the same paper\cite{jeong2011weighted}. It is similar to WDTW, but weight factors are multiplied with differences of derivatives rather than differences of magnitude.

Constraint-based approaches are used to restrict the skewness of the DTW warping path as well as to increase the runtime speed. The most famous and widely used approach is windowing\cite{salvador2007toward} or lower bounding the distance. Windowing restricts DTW warping path to remain only inside the window created in the distance matrix. The shape of this window can vary depending on the domain knowledge or fluctuations in the time-series data. This approach helps improve both computation time and accuracy if the window is selected carefully.

Attempts are also made to use DTW in Machine Learning applications. For instance, DTW distance is being used as Neural Network node value \cite{cai2019dtwnet} in the classification of time-series data, which is referred to as DTWNet. DTWNet is a simple neural network with one or more layers as DTW layer/s. Each DTW layer's nodes optimize the DTW distance value via a back propagation. In another attempt, DTW loss function is made differentiable in order to use it for regression. In SoftDTW \cite{cuturi2017soft} the DTW loss function is differentiable, and hence, allows the computation of its value and gradient in quadratic time and space.

\subsection{Related Work}
\label{sec:relatedwork}

A large amount of time-series data are being generated every day in many fields, such as manufacturing, oil and gas industries, engineering, finance, medicine, natural science and biology, etc. Due to this, many interesting applications of DTW are seen in data mining. For example, DTW is being applied to fuzzy clustering \cite{izakian2015fuzzy}, clustering with global averaging method on DTW \cite{petitjean2011global}, clustering with hidden Markov model and DTW \cite{oates1999clustering}, WDTW for time-series data classification \cite{jeong2011weighted}, support vector based algorithm with WDTW for time-series data classification \cite{jeong2015support}, and motif discovery \cite{mueen2014time}. It has been extensively used in speech recognition \cite{sakoe1978dynamic, godin1989dtw}, handwriting recognition \cite{bahlmann2002online}, gesture recognition \cite{campbell1996invariant}, signature recognition \cite{faundez2007line}, ECG signal pattern recognition \cite{huang2002ecg}, and many others.

Due to many important applications of time-series data classification, many DTW variations are proposed. 
Standard DTW is used for image matching by Rath and Manmatha \cite{rath2003word} and its performance is compared with other popular techniques. Gullo \cite{gullo2009time} proposed time-series data representational model, called derivative time-series segment approximation. Jeong\cite{jeong2011weighted} applied weighted DTW (WDTW), Derivative DTW (DDTW) along with weighted derivative DTW (WDDTW) on synthetic as well as real-life datasets like Swedish leaf, lightening-2, ECG etc. and compared their performance with other techniques.

Keogh and Pazzani proposed derivative DTW (DDTW)\cite{keogh2001derivative}. Cleverly, they did not consider the actual distance between the magnitudes. Instead, they considered the difference between derivatives to prevent pathological warpings. Their work used only three data sets. They did not test it in the context of the classification of time-series data. Later works, however, used DDTW and performed the evaluation \cite{jeong2011weighted, fu2011review}. 
On the basis of this method,
Kulbacki created a measure that takes into account the standard distance
between time-series\cite{kulbacki2002unsupervised}. To combine Euclidean distance measure and estimated derivative distance, they considered the product of these two.

Weighted dynamic time warping (WDTW) is being used for speech recognition \cite{zhang2014one}. It is also used for satellite image time-series analysis \cite{wegner2019dtwsat}. Furthermore, it has been extended to time-weighted DTW \cite{maus2016time}.


DTW is used to find dissimilar action behaviors of online game players \cite{thawonmas2008visualization} in order to visualize them. It is being used to align the sagittal image sequences in dynamic visualization of vocal tract shapes during speech in 3D \cite{zhu2012dynamic}. DTW is also used to find the patterns in human motions \cite{hachaj2017advanced}. A new algorithm SUBDTW is proposed \cite{lee2009visualization}, which is based on DTW, to visualize local temporal trends in multivariate time-series data. 

Even after all these works, there are no proper guidelines for choosing a variant of DTW for a specific problem domain or for signals with particular characteristics. In our work, we try to fill this gap by evaluating these different variants over different characteristics of time-series sequences and present the results as a guideline.


%% file: Content/SynthesisFramework.tex
\section{A Synthesis Framework for Time Series Data}
\label{sec:synthesis}

\begin{figure}[htb]
    \centering
    \includegraphics[width=1.0\linewidth]{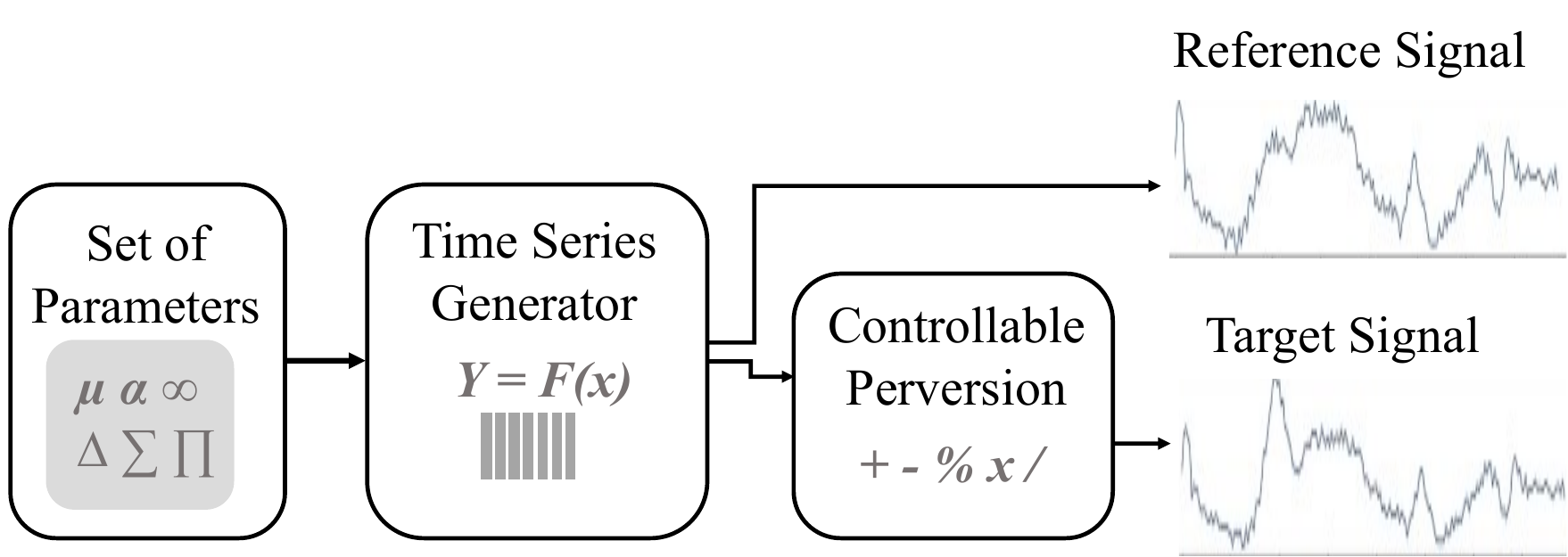}
    \caption{Our time-series synthesis framework consists of two steps. The first step produces an initial series, while the second step deforms the initial series with the controllable variations to generate the second series.}
    \label{fig:SystemArchitecture}
\end{figure}

There are many possible variations between two time-series data (e.g., two 1D signals) processed by a matching or alignment algorithm. It is difficult to have all such different variations in a single pair of real-life time-series data and knowing the variations between them may help us better understand how these variations impact the performance of different DTW measures in different tasks. This leads us to create a synthesis framework capable of generating pairs of time-series data with the desired features and variations.
In this section, we will describe how our synthesis framework generates realistic time-series sequences and series pairs with controllable variations for the subsequent evaluation. We also assess the effectiveness of our synthesis framework by applying it to fit a few real-world signals.

\subsection{Realistic 1D Signal Generation Technique}
\label{sec:ourframework}

A linear temporal signal can have different features. In this work, three of them are considered, that is, the range of its magnitude, the distance between two successive peaks/valleys, and the nature of function it follows between peaks/valleys. Considering these three main features, a synthesis framework capable of creating realistic time-series sequences is developed.

\noindent \textbf{Input parameters:} Our synthesis framework provides users the control over the range of magnitude of the signals (Min, Max), the allowed range of distance between two successive peaks/valleys ($p_1$, $p_2$), length of the signal (L) and the list of allowed functions which can be used to interpolate the signal between successive peaks/valleys. In the experiments shown in this work, functions of the form $y = cx^a + dx^b$ are used where $a$ and $b$ vary between 0.1 and 2 randomly and ($c,d$) are coefficients fitted according to the locations of the two endpoints (peaks/valleys).

\begin{figure}[th]
    \centering
    \includegraphics[width=1.0\linewidth]{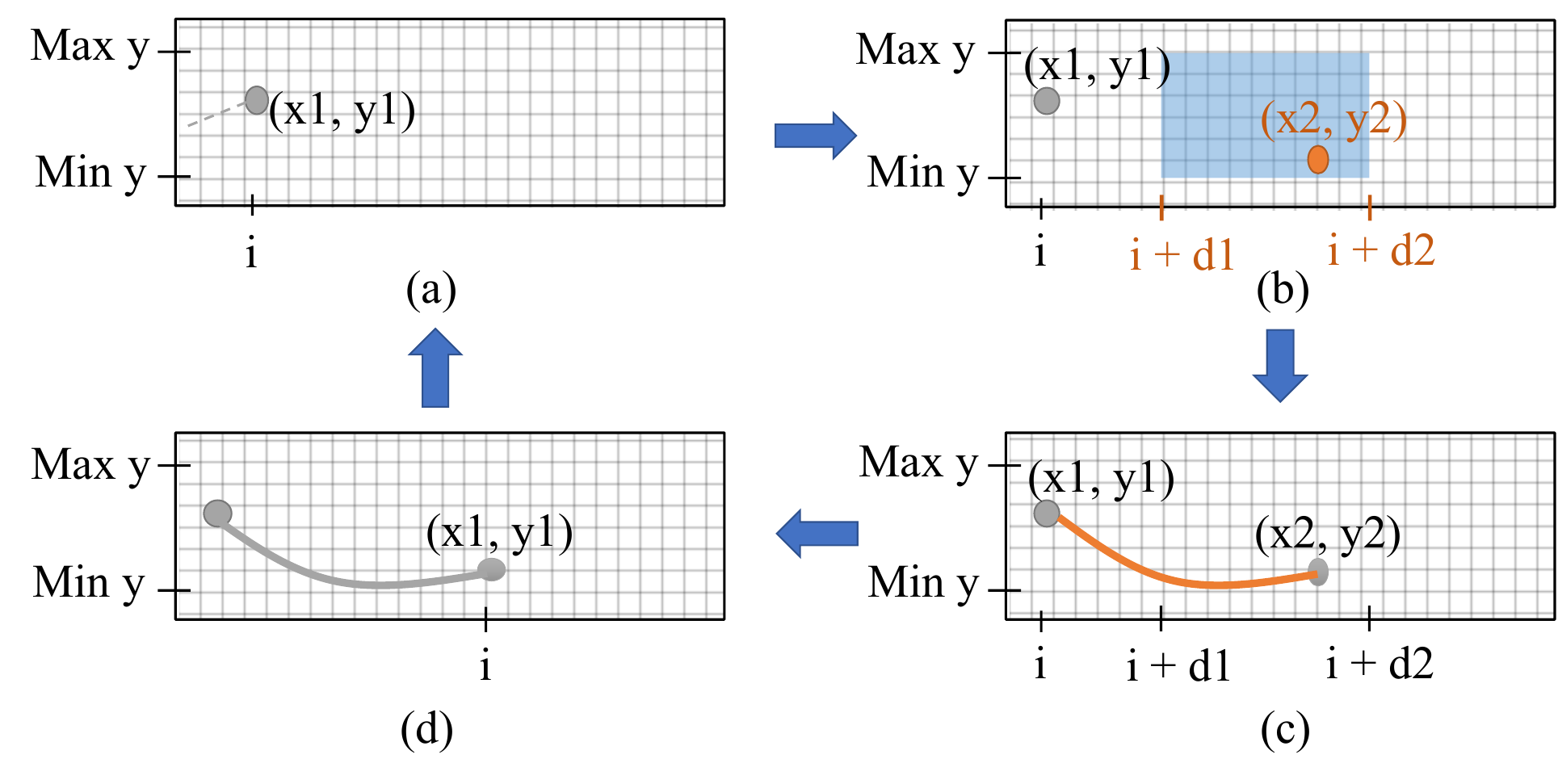}
    \caption{Time-Series Sequence Generator. (a) Sample a point between minimum and maximum magnitude; (b) Randomly sample another point between minimum and maximum magnitude and place it at random distance between maximum and minimum allowed distance between two peaks/valleys from first point; (c) Generate a random function like $y = c_1x^a + c_2x^b$ and fit it between these two points; (d) Add these sampled points to the signal array and repeat this process until the desired length is reached.}
    \label{fig:generateinitialseries}
\end{figure}

\begin{figure}[htp]
    \centering
    \includegraphics[width=1.0\linewidth]{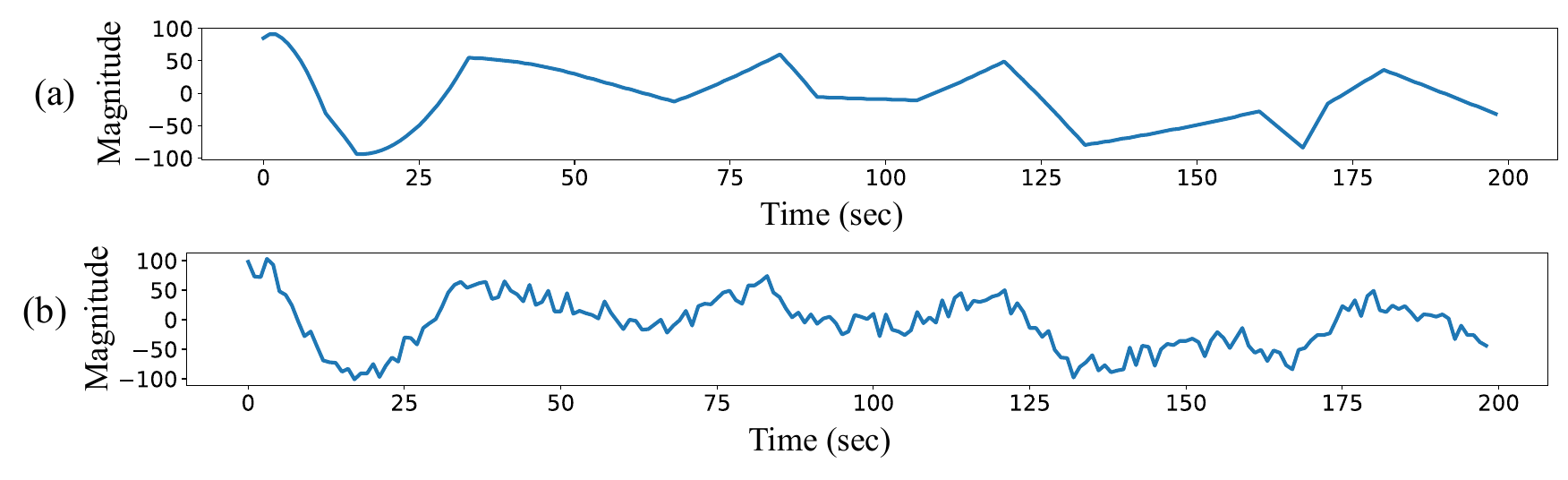}
    \caption{Initial synthetic signal generation without (a) and with noise (b) inserted. }
    \label{fig:genInitial}
\end{figure}

\noindent \textbf{Generation:}
After getting these specifications from the user, our framework generates a signal which may not look realistic initially (Figure \ref{fig:genInitial}(a)), but then it adds random noise with $+-20$\% of the magnitude range (Max-Min) (Figure \ref{fig:genInitial}(b)). In particular, our framework creates a point $(X_1, Y_1)$ in the range of (Min, Max), it then creates a rectangular region spanning the range of (Min, Max) and a range along time axis, ($p_1$, $p_2$), as shown with blue shade in Figure \ref{fig:generateinitialseries} (b) to sample another point $(X_2, Y2)$. Next, it randomly selects a function from the given set of functions to fit the signal between $(X_1, Y_1)$ and $(X_2, Y_2)$. This process is repeated until the desired length of the sequence is reached. Algorithm-\ref{alg:initialsignalgen} and Figure \ref{fig:generateinitialseries} illustrate this process.

\begin{algorithm}[th]
\textbf{Input:} Length of Signal (L), Range of Magnitude for Sampling (Min, Max), Min and Max distance between two peaks/valleys ($P_1$, $P_2$).\\
\textbf{Output:} \text{Synthetic Signal.}
\begin{algorithmic}[1]
\STATE Initialize an empty list S
\STATE PrevY = pick a random integer between Min and Max
\STATE PrevX = end = 1
\STATE Append PreY to S
\WHILE{$end < L$}
\STATE $RangeStart = PrevX + P_1$
\STATE $RangeEnd = PrevX + P_2$
\IF{$RangeStart \geq L$}
\STATE Stop and return S
\ENDIF
\IF{$RangeEnd \geq L$}
\STATE $RangeEnd = L - 1$
\ENDIF
\STATE x = a random integer between RangeStart and RangeEnd
\STATE y = a random integer between Min and Max
\STATE Pick two numbers a and b in [0.1, 3] and create functions as $y = c_1x^a + c_2x^b$ and $PrevY = c_1PrevX^a + c_2PrevX^b$
\STATE Solve the above equations for $c_1$ and $c_2$
\STATE Fit the function $y = c_1x^a + c_2x^b$ between PrevX and x and append the y values to S
\STATE PrevX = x
\STATE PrevY = y
\ENDWHILE

\STATE Initialize empty list N
\FOR{i in range 0 to Len(S)}
\STATE $t = randomInt(0, 0.4*Max) - 0.2*Max$
\STATE append t to N
\ENDFOR
\STATE S = S + N
\STATE Return S
\end{algorithmic}
\caption{ Algorithm $Generate Signal$. 
}
\label{alg:initialsignalgen}
\end{algorithm}

\subsection{Controllable Distortion to Reference Signal}

Next, the framework uses the initial signal generated above as a reference and produces another signal (target) by perverting (or modifying) this reference signal. 
The overall pipeline of this framework is shown in Figure~\ref{fig:SystemArchitecture}. There are two basic deformations/features which can be composed to create other variations. Using these basics and composite variations any temporal signal can be transformed to any other temporal signal, which will be detailed in the later sections.

\subsubsection{\textbf{Basic Modification 1:} Scaling}
\label{sec:modification1}

The first basic modification we consider is the scaling. This scaling will modify the length of a portion of the reference uniformly. This scaling may change the length of the target (or output) signal compared to the reference. We provide the option for the user to decide whether they want to ensure the identical length between the target and reference signals or not. 

\noindent\textbf{Input parameters:} The scaling takes the beginning of the window for scaling, $W_0$, and its ending, $W_1$, as well as a scaling factor $s$ (usually ranging from $0.5$ to $1.5$) as the input.

\noindent \textbf{Scale a Random Portion of the Signal.}
Let the target signal created after the controlled deformation (scaling) be y and the reference signal be x when the window from index $W_0$ to index $W_1$ is scaled with the scaling factor of $s$. Then the magnitude of the signal at $j^th$ index of y is taken from the $i^th$ index of x where $i$ is determined using the following formula (rounded to the nearest integer).

\begin{equation}
  i=\begin{cases}
    j, & \text{if $j<W_0$ }.\\
    W_0+(j-W_0)s, & \text{if $W_0 \leq j \leq W_0+(W_1-W_0)s $}. \\
    (W_1-W_0)(1-s) + j, & \text{otherwise}.
  \end{cases}
\end{equation}

\begin{figure}[htp]
    \centering
    \includegraphics[width=1.0\linewidth]{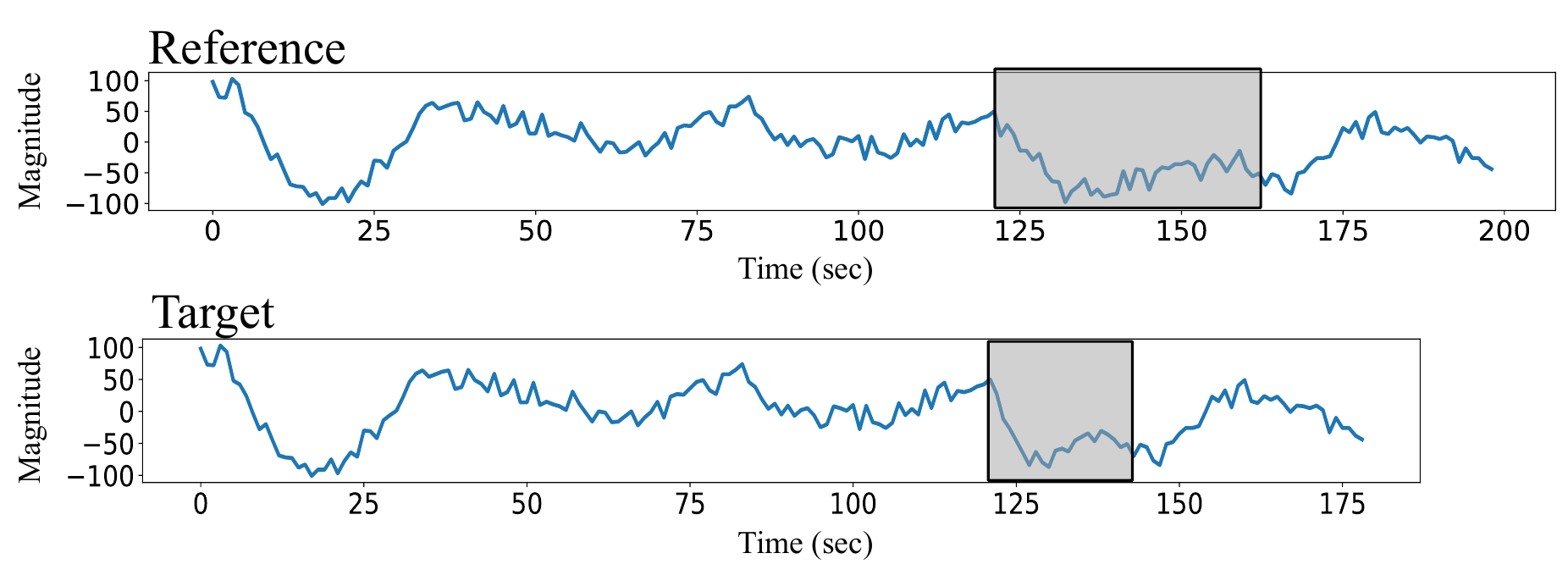}
    \caption{Scaling the shaded region without length preservation.}
    \label{fig:Scaling}
\end{figure}

Here, the window from index $W_0$ to $W_1$ on the reference signal is scaled by the scaling factor ($s$), and the later section of the sequence is shifted to accommodate this scaling.
Figure \ref{fig:Scaling} shows this variation between the sequences. Gray boxes show the scaling window in both the reference and target signals. In this particular instance, the window is scaled down to 51\%.

\noindent\textbf{Scale a Random Portion While Preserving Signal Length.}
Time-series data sequences often have their features shifted even if they represent similar physical quantity and have the same length. That said, if a portion of the reference is scaled, other portions also need to be scaled and shifted to preserve the length. To achieve that, let the target signal created after controlled deformation on reference signal x be y. When the window from index $W_0$ to $W_1$ is scaled with the scaling factor of $s$ and remaining sections with $s'$ to keep the size of the target signal the same as the reference. Let the length of the original signal be L, and the length of the scaling window ($W_0$, $W_1$) be $W$, then $s'$ can be computed as $s' = \frac{L - W\times s}{L - W}$. The magnitude of series at $j^{th}$ index of y is taken from the $i^{th}$ index of x where $i$ is determined as follows (rounded to the nearest integer).

\begin{equation}
  i=\begin{cases}
    j s', & \text{if $j \leq W_0 s'$ }.\\
    \frac{W_0 s'(s - 1) + j}{s}, & \text{if $W_0 s' < j < W_0 s' + (W_1-W_0)s $}. \\
    \frac{(W_0s' + (W_1-W_0)s)(s'-1) + j}{s'}, & \text{otherwise}.
  \end{cases}
\end{equation}

\begin{figure}[htp]
    \centering
    \includegraphics[width=1.0\linewidth]{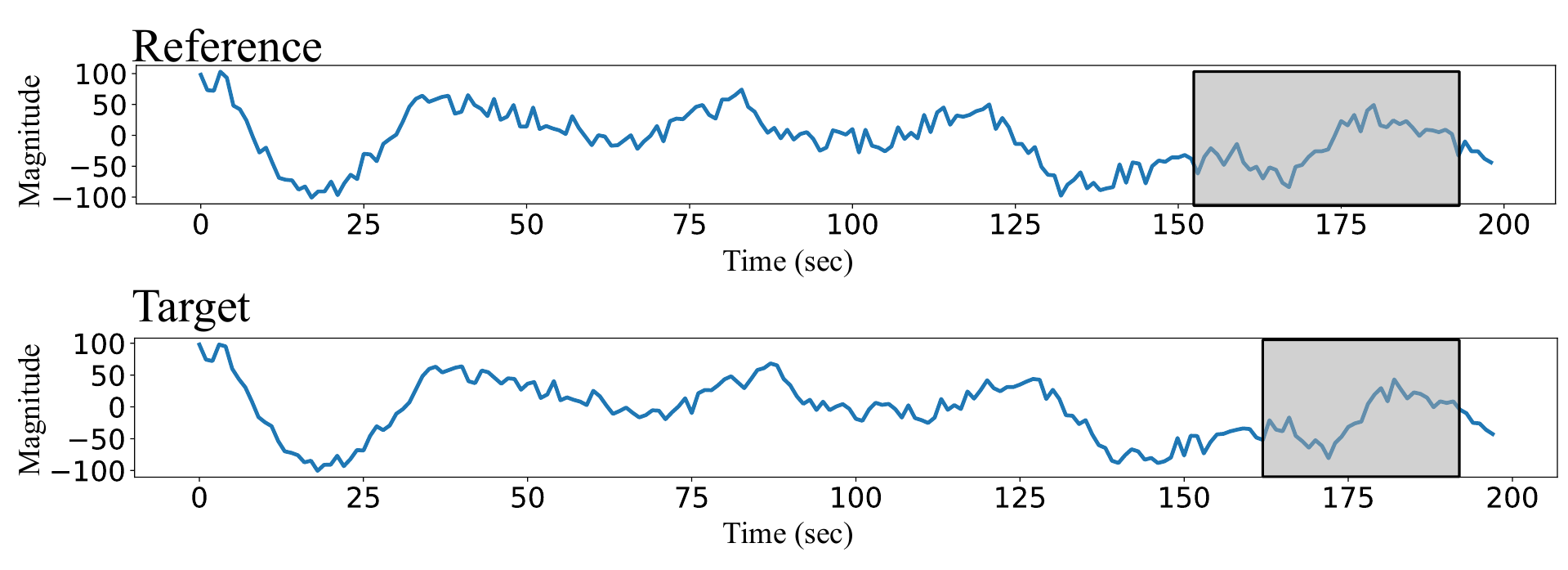}
    \caption{Length preserving scaling. Adjustment is made in the shaded region.}
    \label{fig:FullScaling}
\end{figure}

Figure \ref{fig:FullScaling} shows such an example of scaling with the preserved length. The gray boxes highlight the portion of the signals that were adjusted after the initial scaling shown in Figure \ref{fig:Scaling}.

\subsubsection{\textbf{Basic Modifications-2}: Addition of Random Gaussian Peaks to The Time-Series}
\label{sec:modification2}

\begin{figure}[htp]
    \centering
    \includegraphics[width=1.0\linewidth]{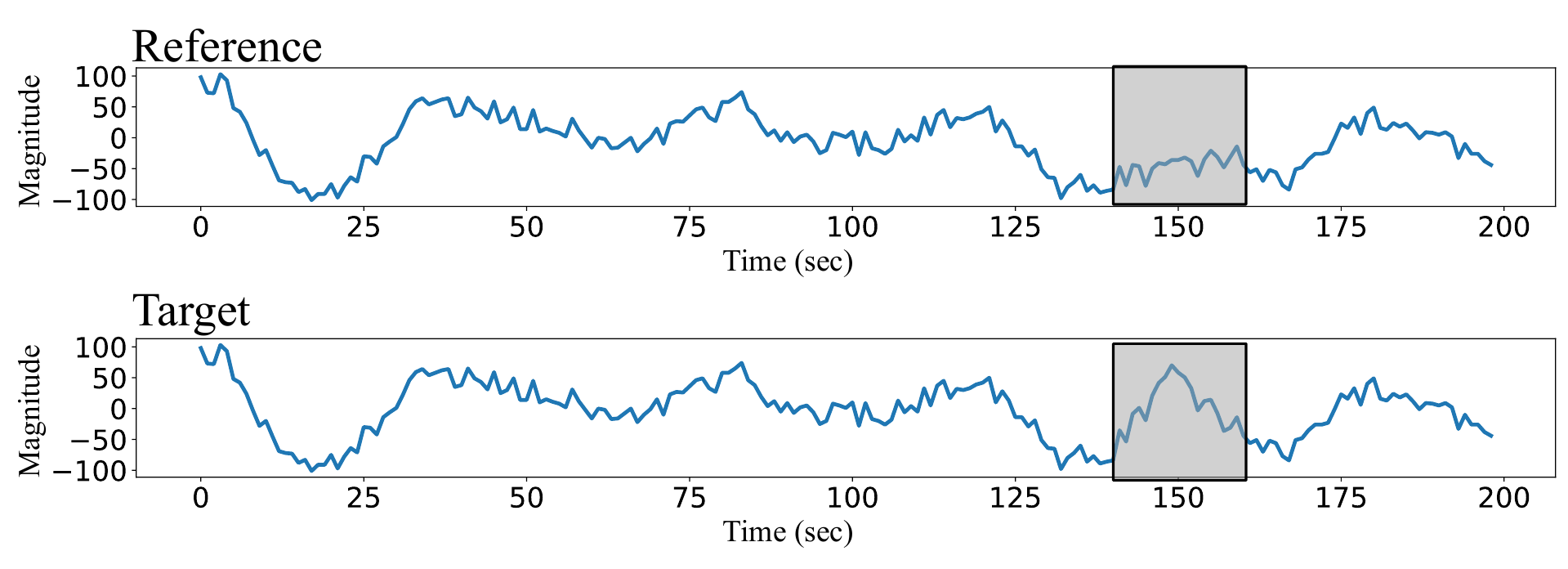}
    \caption{Adding a random Gaussian peak in the shaded portion.}
    \label{fig:RGP}
\end{figure}

Sometimes, two time-series sequences have similar features at similar locations on the time axis, but they may have some extra features or features missing from one of them. This variation is sometimes due to noise or human errors.
Simply increasing or decreasing the magnitude at a single point to achieve the above variation is usually not sufficient, as it may be equivalent to inserting a random noise, making the resulting signal unrealistic.
To address that, our framework adds a Gaussian peak. Specifically, it increases the magnitude of the signal within a small window with its center having the highest magnitude and gradually decreasing away from the center. Figure \ref{fig:RGP} provides such an example. 

\noindent\textbf{Input parameters:} This modification takes the position of the center $i$ and the width $n$ of the Gaussian, as well as the magnitude $h$ at the center as the input. The Gaussian function is then computed as $g(j)=h\exp{(-(\frac{j-i}{2n})^2})$.

Let us assume the Gaussian peaks are $\{ g_1, g_2, g_3...g_n\}$ and we add this to the signal y at index i, then the index window where this peak will be added is $(W_0, W_1)$ $W_0 = i - \frac{n}{2}$ and $W_1 = i + \frac{n}{2}$.
\begin{equation}
  y_j=\begin{cases}
    x_j, & \text{if $j<W_0$ or $j>W_1$ }.\\
    x_j + g_{(j-W_0)} & \text{Otherwise}. \\
  \end{cases}
\end{equation}



Instead of just one peak, we can also add multiple random Gaussian peaks \textbf{(MRGP)} $\{ g_1, g_2, g_3...g_n\}$ at random places on the signal. This will create a signal with more variation.

Note that subtracting an RGP can be achieved by adding an RGP with a negative magnitude.

\subsubsection{\textbf{Composite Modification :} Scaling along with Addition of Gaussian Peaks}
\label{sec:modification4}

This type of variation is widely observed in real-life time-series data. Usually, two similar time-series data sequences measuring similar events (e.g., drilling well signals) have a shifting of features and some additional features, or some of the features are missing from one of them. This type of variation can be considered as the combination of the previously described two variations (i.e., shifting and addition/subtraction of peaks).

To achieve the combination of shifting and RGP insertion or removal, first, the reference signal is scaled similarly to \textbf{modification 1}. Then a Gaussian peak is added at a random place on this scaled signal similar to \textbf{modification 2}. This new perverted signal has much more variation compared to those produced by the previous modifications, since it changes both the temporal property by scaling and also local features by the addition of peak(s).
The mathematical formula is similar to the one described in Section \ref{sec:modification2}, but this time instead of adding it to the reference, we add the peak to the scaled signal.

\begin{figure}[htp]
    \centering
    \includegraphics[width=1.0\linewidth]{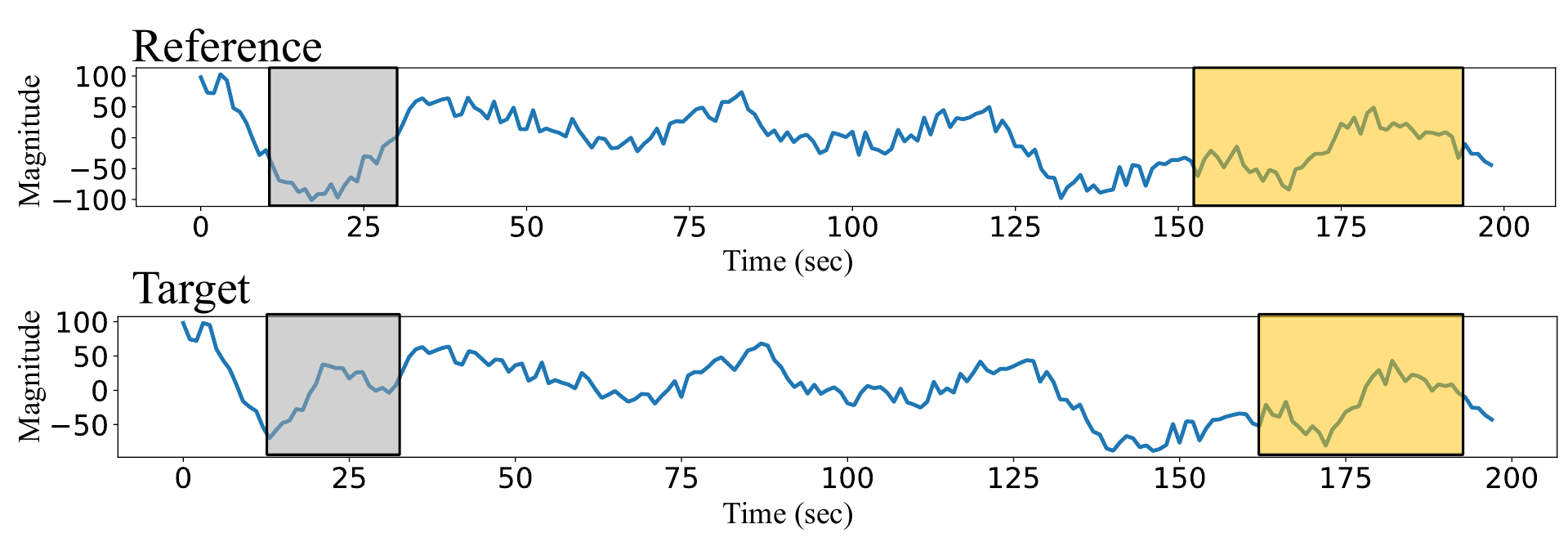}
    \caption{Modification by both scaling (in the yellow shaded region) and adding a random Gaussian peak (in the gray shaded region).}
    \label{fig:SRGP}
\end{figure}


\subsection{Effectiveness of Our Synthesis Framework}
\label{sec:simulatedannealing}
To prove that any temporal signal can be transformed to another temporal signal using only the above modifications, a simulated annealing \cite{van1987simulated} based framework is developed to optimize the parameters used by our synthesis framework in order to enable the generation of a synthetic signal that is sufficiently close to the real one.  

Take two signals as input. Without loss of generality, it considers one as the source and the other as the target. The goal is to transform the source signal to make it look as close as possible to the target by performing only the above modifications. It uses Euclidean distance as a measure for evaluating how close two temporal signals are. Basically, simulated annealing uses Euclidean distance between the source and generated target with the current parameters as the objective function and tries to minimize this function.

There are two steps in our fitting framework. First, scaling is performed on the source signal using simulated annealing. It is mainly used to resize the source in order to match the length of the target signal. There are three parameters to fit in this process, location of scaling, width of the scaling window, and the scaling factor (magnitude of scaling). To simplify and speed up this search, the width of the scaling window is set to 50\% the original length of the source. Let the width be W,  then 
$W = 0.5 \times length(source)$. Another parameter, the scaling factor, is computed using the following formula
$$s = 1 + \frac{length(target) - length(source)}{W}$$

The third parameter, i.e., the location of this window, is found via simulated annealing by minimizing the Euclidean distance between the source and the target. Once the source is scaled and its length matches the length of the target, it is passed to the second step.

\begin{figure}[tb]
    \centering
    \includegraphics[width=1.0\linewidth]{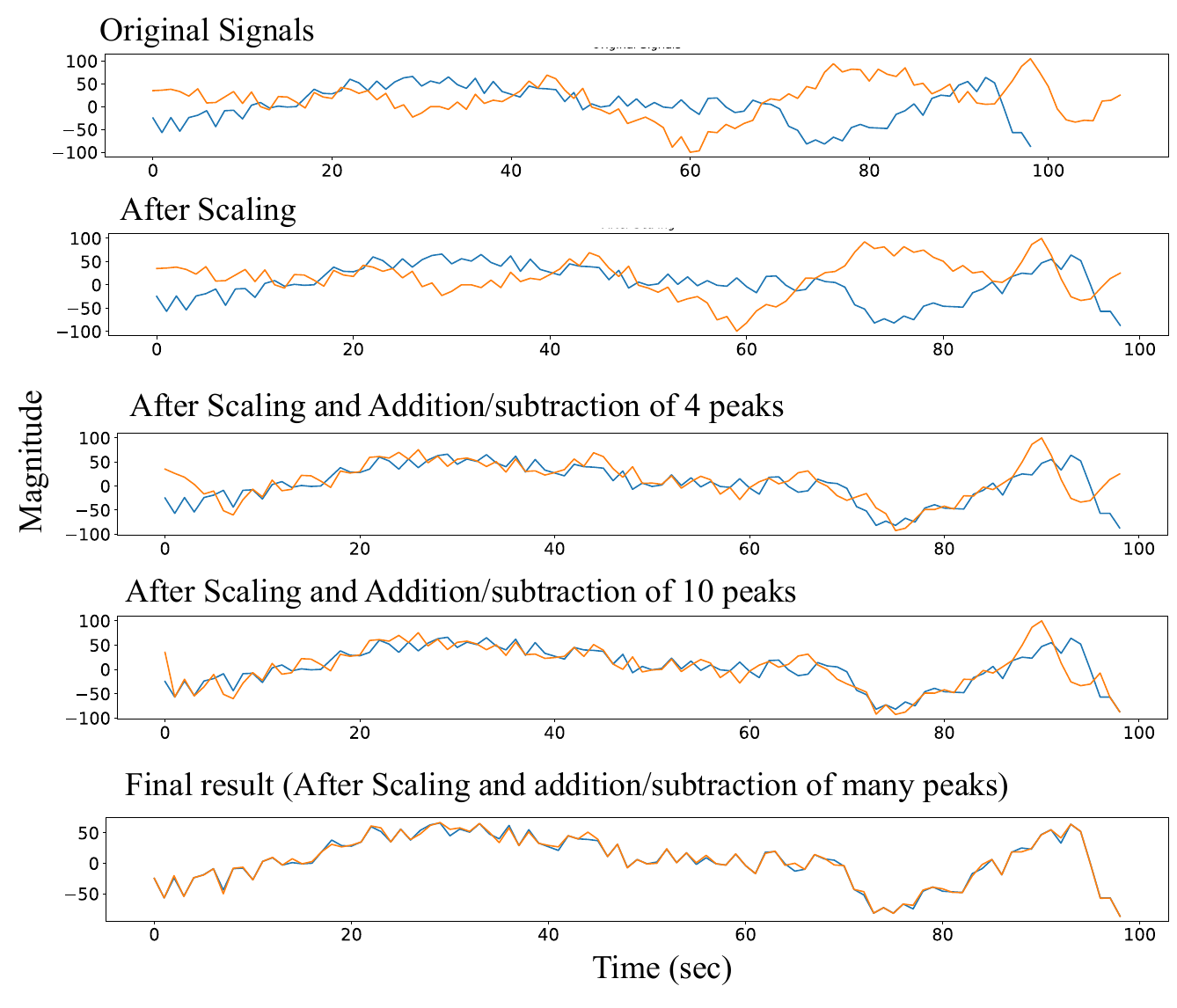}
    \includegraphics[width=0.8\linewidth]{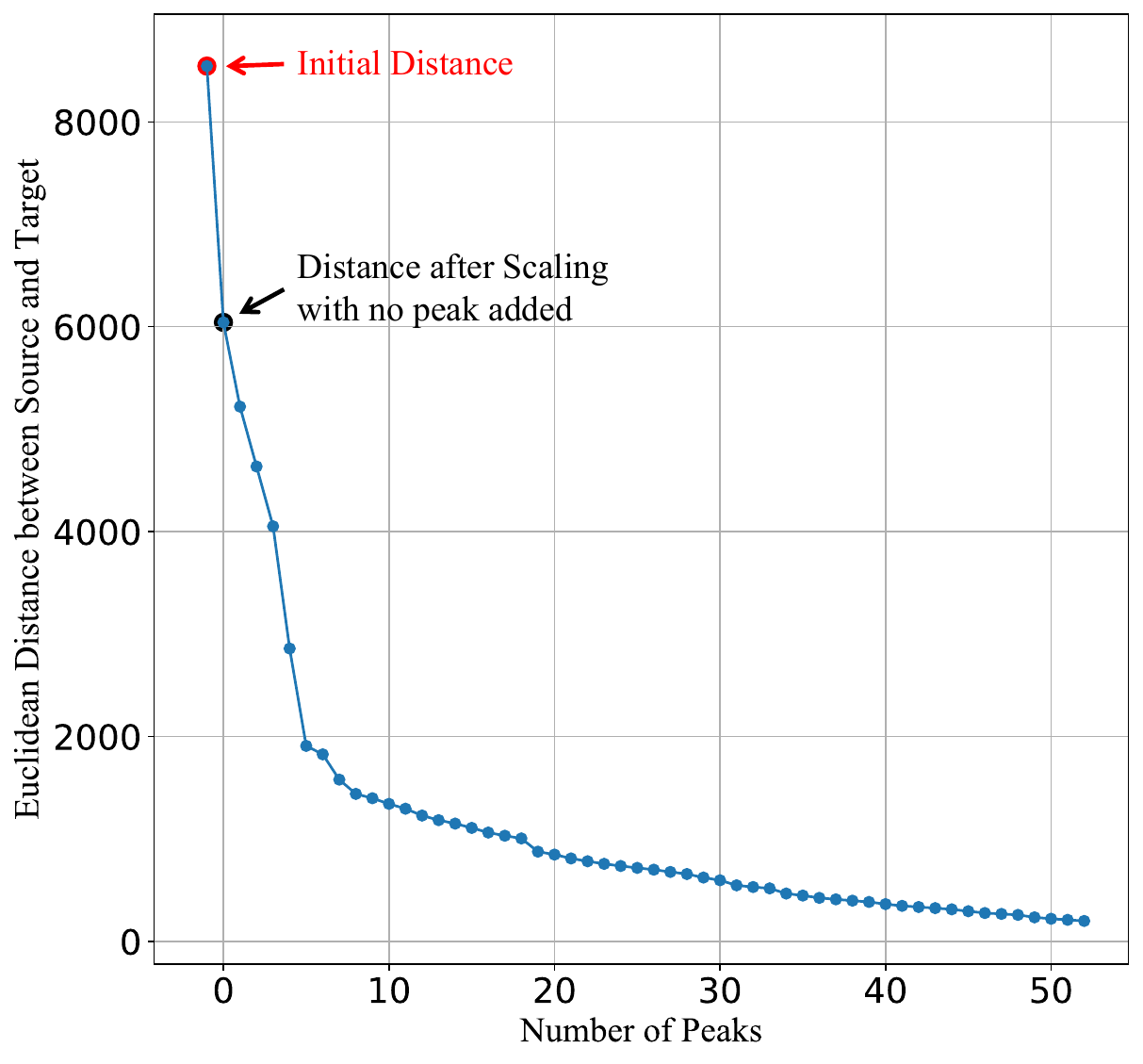}
    \caption{Signal-fitting result. The top row shows the source and the bottom row shows the fitted result. Middle rows show the intermediate fitting results. The last plot shows the Euclidean distance between Source and Target time-series through the simulated annealing iterations.
    }
    \label{fig:SA_Result1}
\end{figure}


In the second step, multiple Gaussian peaks are added to the source with the help of simulated annealing. Simulated annealing is performed multiple times to reduce the Euclidean distance between the two signals. In each iteration, one peak is added. There are also three parameters to optimize, including the location of the peak, the width of the peak, and the magnitude of the center of the peak. All three parameters are initialized randomly and optimized by simulated annealing for each Gaussian peak. Multiple peaks are required to be added in order to match the curves of the target. It keeps adding the peaks until the Euclidean distance between the source and the target is smaller than a threshold (T). The threshold is proportional to the length and magnitude range of the target signal as shown below.
$$T = \frac{x}{100} \times magnitude(target) \times length(target)$$
Here $x$ can be used to control the accuracy of fitting. 

We performed experiments to evaluate the above signal-fitting framework. In these experiments, we fixed, $x=1$, i.e., 1\% error is allowed. Figure \ref{fig:SA_Result1} shows how the fitting process using the proposed synthesis framework can deform the source signal and make it become the target signal.


There is a trade-off between the number of peaks added/subtracted and the error between source and target. In a noisy signal pair, while transforming the source to look like the target, a few initial peaks after scaling transform the overall trend of the source to be sufficiently close to the target.
Once large peaks are adjusted, the framework starts adjusting small peaks. After two signals almost have an identical trend, it tries to adjust an individual (or two adjacent) point/s to make the source look exactly like the target. Allowing large numbers of peaks to be added/subtracted by the framework, it is capable of completely transforming a source's linear signal to look exactly like the target linear signal. But in general, we do not wish to use a large number of peaks to deform the source to the target. In our experiment, we found the number of peaks needed to produce a signal with 1\% distance from the target is at most half of the number of samples on the source.

\begin{figure*}[!ht]
    \centering
    \includegraphics[width=1.0\linewidth]{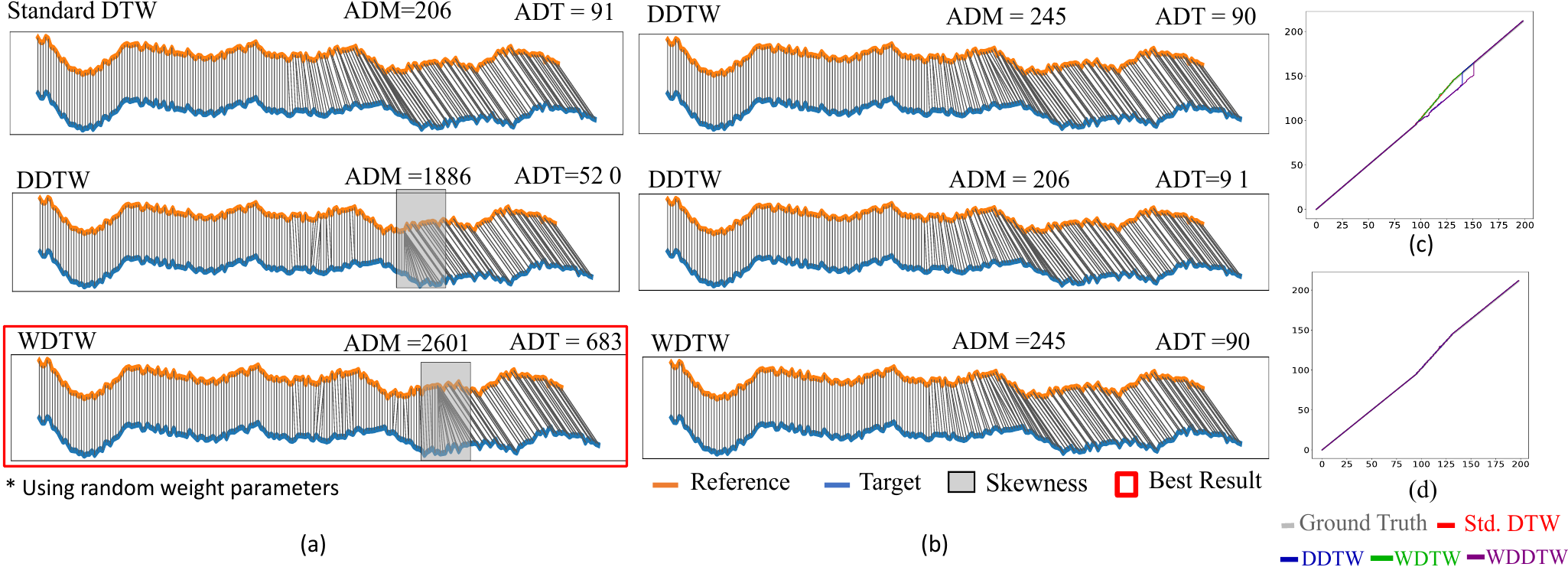}
    \caption{
     \label{fig:Result_Scaling}
    Results from different variants of DTW on signal pairs where the target is simply scaled on the time axis at certain places. (a) Weight parameter\cite{jeong2011weighted} (g = 0.4, 0.4) of WDTW and WDDTW are simply selected at random. (b) The weight parameter (g=0.21, 0.11) is optimized using Monte Carlo sampling method for both WDTW and WDDTW.
    (c) Warping path for \textit{(a)} with randomly selected weight parameter (g) both WDTW and WDDTW deviates from ground truth at the central region while both DTW and DDTW mostly follow the ground truth path with some staircase effect to accommodate scaling. (d) Warping path for \textit{(b)} With optimized weight parameter (g) WDTW and WDDTW along with standard methods mostly follow the ground truth path.
    }
    \vspace*{\floatsep}
        
    \includegraphics[width=1.0\linewidth]{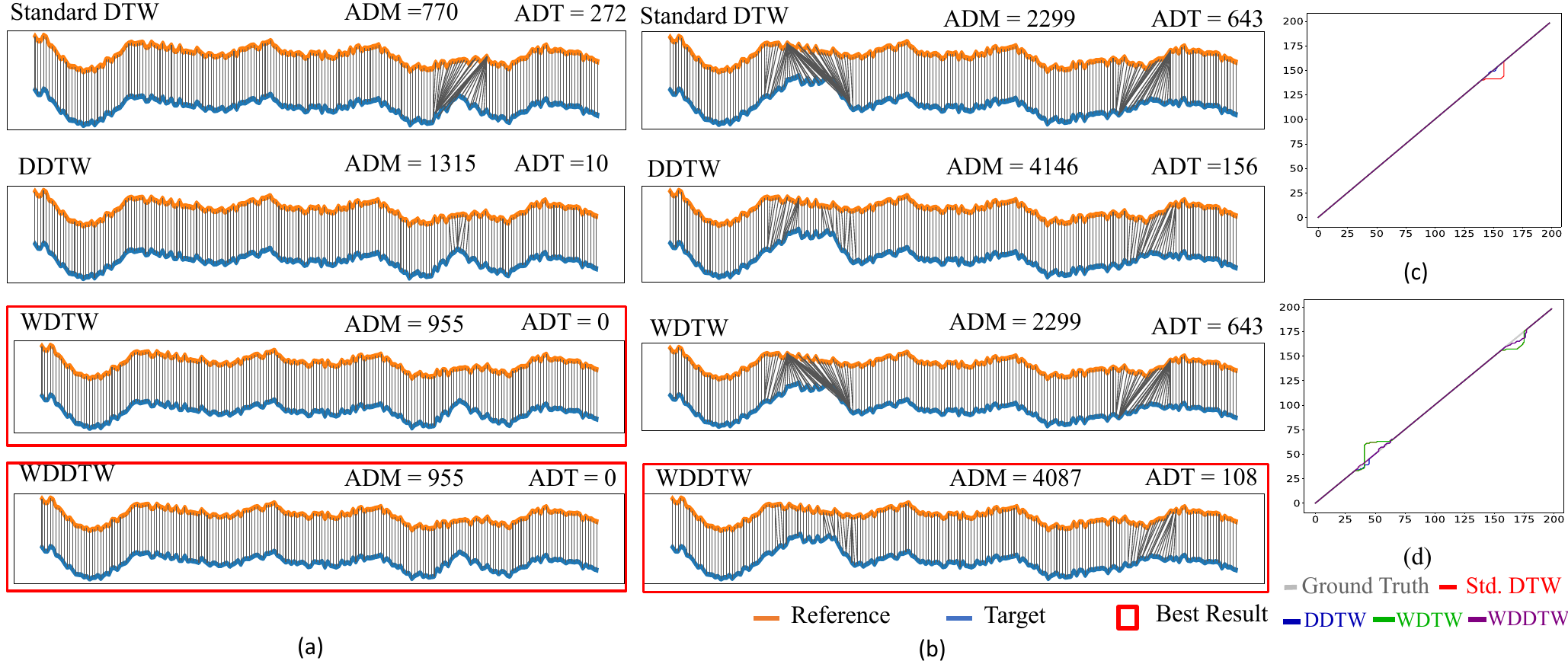}
    \caption{
        \label{fig:Result_RGP}
        Results from DTW variants when the target has a random Gaussian peak added. (a) Target has only one random Gaussian peak (RGP), with the ADM measure, Standard DTW performs the best while based on the ADT measure, Weighted DTW variant(s) with optimal weight parameter outperforms DTW and DDTW. (b) Target has multiple RGP, Derivative variants (DDTW and WDDTW) perform well on ADT measure while Standard DTW methods work the best considering the ADM measure.
    (c) Warping path for \textit{(a)} with single RGP Non-weighted variants jiggles a little around the peak. (b) Warping path for \textit{(b)} with multiple peaks Weighted DTW variants could not accommodate all the peak adjustments since weights are optimized based on ADM measure, they try to minimize the magnitude difference and end up with those deviations from the ground truth path.
    }
\end{figure*}

%% file: Content/Evaluation.tex
\section{Evaluation of DTWs with Synthetic Time-Series}
\label{sec:evaluationofDTW}

With the above synthesis framework, we can generate sets of synthetic signals with controllable and known variations. Using them, we perform the evaluation on the performance of different DTW measures on two time-series data analysis tasks, i.e., alignment and classification. We specifically focus on the alignment task, as it is required for many other tasks.

\subsection{DTWs for Signal Alignment}
\label{sec:alignmenteval}

For this evaluation, we produced sets of synthetic signal pairs with one type of variation (e.g., scaling only, one RGP, combined scaling and RGPs, etc.) and performed signal alignment between the individual pairs using different DTW measures. We quantitatively measure the alignment accuracy given the known correspondence between the samples on the corresponding pairs (based on the index correspondence during the signal deformation) and report the performance of those DTW measures. Both ADM and ADT distance measures are used in our evaluation (Table-\ref{tab:results}). The following discussion is organized based on the type of variations that are imposed on the signal pairs. For each group of experiments, 50 signal pairs with varying parameter values were generated and utilized, and a representative result was selected for each group for the discussion. 

\subsubsection{Scaled Signals}
\label{sec:ScaledResult}
\textbf{When signal features are shifted on the time axis with no variation in magnitude, standard DTW methods (DTW, and DDTW) seem to work the best among these four variants}. In Figure \ref{fig:Result_Scaling} (a) when weight parameter (g) is not optimized, weighted methods usually perform worse than the standard methods. A gray shaded box covers the regions where WDTW and WDDTW try to accommodate the shifting of the features on the time axis, but they fail miserably, as they try to accommodate the entire shifting at just one point, causing inaccurate matching. In Figure \ref{fig:Result_Scaling} (b) the weight parameter (g) is optimized, both WDTW and WDDTW perform almost similarly to DTW and DDTW as indicated by both the ADM and ADT distance measures.
It can also be verified with visuals in Figure \ref{fig:Result_Scaling} (a) and (b) as the matching looks almost similar. Also, Figure \ref{fig:Result_Scaling} (c) and (d) show the warping path indicates the same.  
In the remainder of the results shown in this work, optimized weight parameter (g) is utilized for all different signal pair variants.


\subsubsection{Length Preserving Scaled Signals}
\label{sec:FullScaledResult}
Next, the target signal that has only shifted features on the time axis with no or negligible variation in magnitude and length compared to the reference signal is considered, as shown in Figure \ref{fig:FullScaling}. There is not much difference in ranking the DTW methods for the alignment with both the ADT and ADM distance measures in this type of series variation. 
In particular, for aligning the time-series pairs with this type of variation, DTW tries to warp the scaling over the time axis by matching points at different indexes on two sequences, thus minimizing the total difference between the magnitude values of the matching points. Weighted DTW, with their weight parameter optimized can also accomplish the same level of accuracy, but they could not outperform standard DTW and DDTW. The visualization of the results of this evaluation and a more detailed discussion can be found in the supplemental document.



In summary for the above two evaluations, \textbf{for the alignment problems when we know there's shifting of features on the time axis and not much difference in the magnitude of the sequences, the recommendation is to use standard DTW methods (either DTW or DDTW) and not weighted variants}, because they may perform worse with random weight parameters and optimizing the weight parameter is itself a very costly process. Even after optimizing the weights, they cannot guarantee to perform better than the standard DTW methods.


\begin{figure*}[tph]
    \centering
    \includegraphics[width=1.0\linewidth]{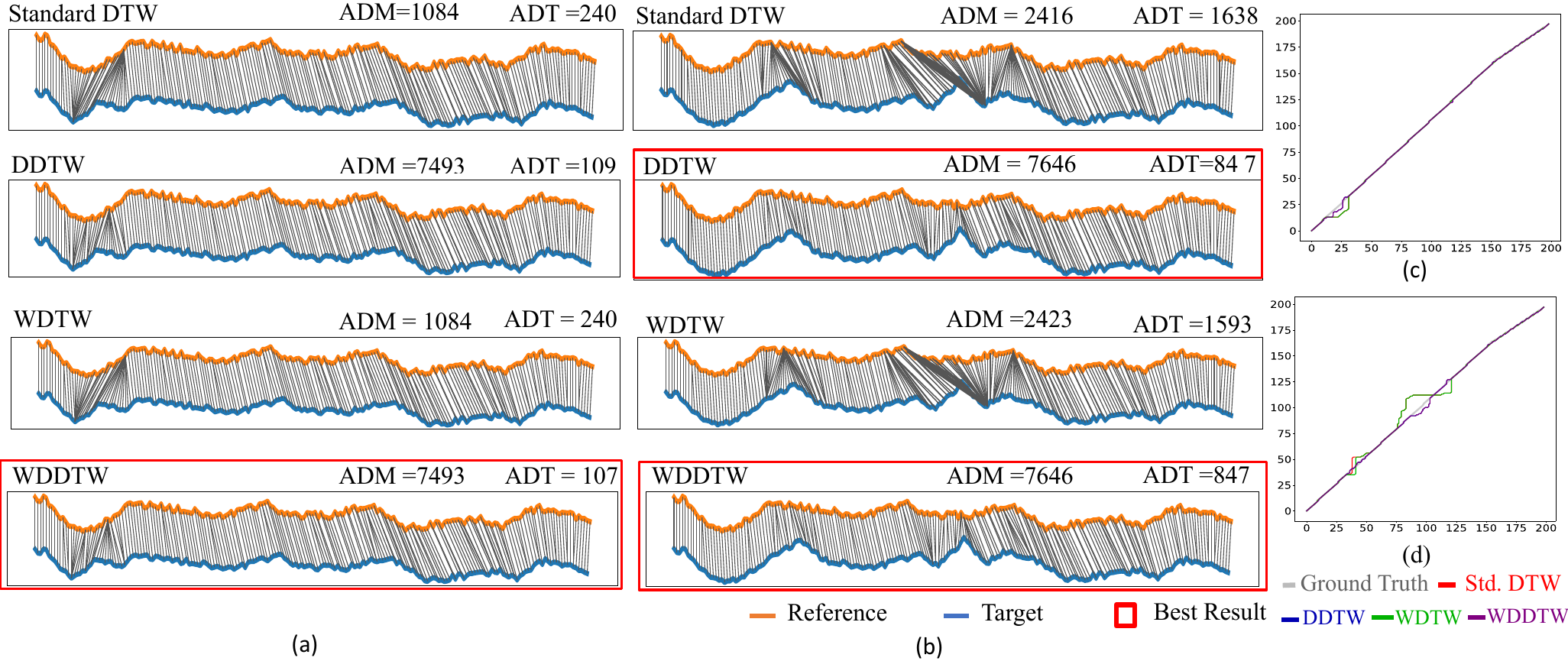}
    \caption{Alignment results from DTW variants when Target contains both Scaling and RGP as described Fig-\ref{fig:SRGP}. (a) Derivative variants of DTW perform well with single peak and scaled signal pairs over ADT measure. (b) WDDTW outperforms others based on ADT over the signal pairs with scaling and multiple RGPs. While Standard DTW is still the best based on ADM measure in both instances.
    (c) Warping path for \textit{(a)} when target series has single RGP with scaling, all four variants fail to align the region around the peak. (d) Warping path for \textit{(b)}, when we have the target scaled with multiple RGP and none of the DTW variants are able to accurately align the pair, but the Weighted variants are slightly closer to the true alignment than the non-weighted variants.}
    \label{fig:Result_SRGP}
\end{figure*}


\subsubsection{Random Gaussian Peak Added}

When two signals have few differences in the features they contain, for example, an extra peak or a missing peak, the widely used distance measure ADM fails. Since it tries to minimize the aggregate difference between the magnitude of matched points, it forcefully matches an extra peak with some point with a high magnitude on the other signal. Different variants of the DTW are also evaluated using this distance measure, since it is widely used in the literature as well as seemly to be the best measure when we do not know the accurate alignment of two sequences. But distance measure in terms of ADT is also provided to evaluate the results given the ground truth alignment. As in Figure \ref{fig:Result_RGP} (a) when the target has one extra peak than the reference does, weighted DTW (with optimized weight parameters) performs better than non-weighted variants (on ADT measure), though this optimization comes at very high time complexity. They could not outperform standard DTW on ADM measure. When the target has differences of one or multiple peaks that \textbf{match the direction of existing peaks} as in Figure \ref{fig:Result_RGP} (b), derivative DTW (DDTW) performs better than standard DTW and WDTW on ADT measure, while they still could not beat standard DTW on ADM measure. Nonetheless, we believe it is not fair to evaluate the alignment based on ADM, specifically for this type of variations.

Figure \ref{fig:Result_RGP} (c) and (d) show the warping paths for the alignment results shown in Figure \ref{fig:Result_RGP} (a) and (b), respectively, which better demonstrate the ADT measure. With an extra peak (i.e., a new peak instead of increasing the magnitude of one of the existing peaks)
, the warping paths of standard DTW and DDTW fail to align the region near the additional peak, while WDTW and WDDTW accurately align them. When one or more additional peaks are added on top of the existing peaks 
as in Figure-\ref{fig:Result_RGP} (c) and (d), the warping paths for standard and WDTW are similar to the paths that DDTW and WDDTW follow. They all fail to correctly align regions near the additional peaks, but DDTW and WDDTW are more accurate because they consider directional information in terms of derivatives.


\subsubsection{Random Gaussian Peak Added on Scaled Signal}
\label{sec:evl_peaksonscaling}

With this type of variation, which is mostly found in real-world time-series data, derivative DTW seems to outperform all others. As shown in Figure \ref{fig:Result_SRGP} (a) and (b), when the target has one or multiple peaks added respectively, WDDTW performs the best among all on ADT measure. But surprisingly, standard DTW is still unbeatable on ADM measure. Figure \ref{fig:Result_SRGP} (c) and (d) show their warping paths, indicating the same.

This variation is widely seen in many real-world applications, such as formation transition alignments, and streamline classification. \textbf{Based on our experimental evaluations, we suggest the usage of WDDTW for the alignment when the signal pair has both scaling and random peaks difference.}

\begin{table}[hb]
  \caption{Performance of different DTW measures (average distance of 50 pairs each experiment)
  }
  \label{tab:results}
  \scriptsize%
	\centering%
  \begin{tabu}{%
	r%
	*{7}{c}%
	*{2}{r}%
	}
  \toprule
   Signal Variation     & {Standard DTW}& {DDTW}    & {WDTW}    & {WDDTW}   \\
                        & ADM(ADT)      & ADM(ADT)  & ADM(ADT)  & ADM(ADT)  \\
  \midrule
  Scaled                & 183(93)       & 211(97)   & 860(331)  & 508(446)  \\
  Scaled but same size  & 851(126)      & 857(132)  & 2506(610) & 1495(755) \\
  RGP                   & 707 (223)     & 196(41)   & 1050(4)   & 223(4)    \\
  MRGP                  & 1920(768)     & 494(103)  & 3079(8)   & 563(10)   \\
  Scaled and RGP        & 1408(406)     & 956(178)  & 3377(638) & 1620(769) \\
  Scaled and MRGP       & 2731(1335)    & 1160(909) & 5418(178) & 1850(153) \\
  
  \bottomrule
  \end{tabu}%
\end{table}

\begin{figure}[th]
    \centering
    \includegraphics[width=1.\linewidth]{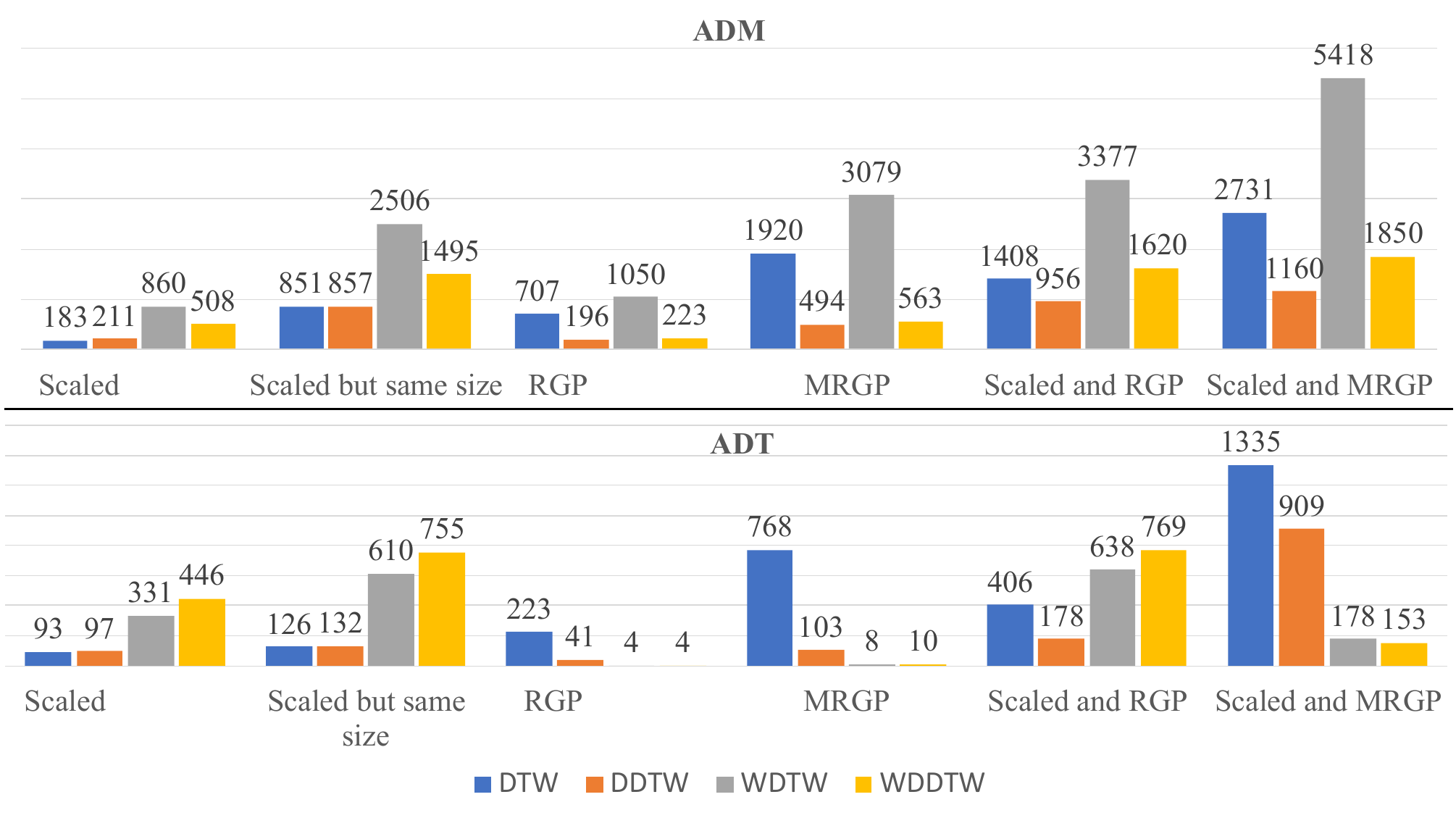}
    \caption{Performance of different variations of DTW. The average distance in 50 signal pairs for each variation is shown.}
    \label{fig:Result_hist}
\end{figure}

\begin{figure}[th]
    \centering
    \includegraphics[width=1.\linewidth]{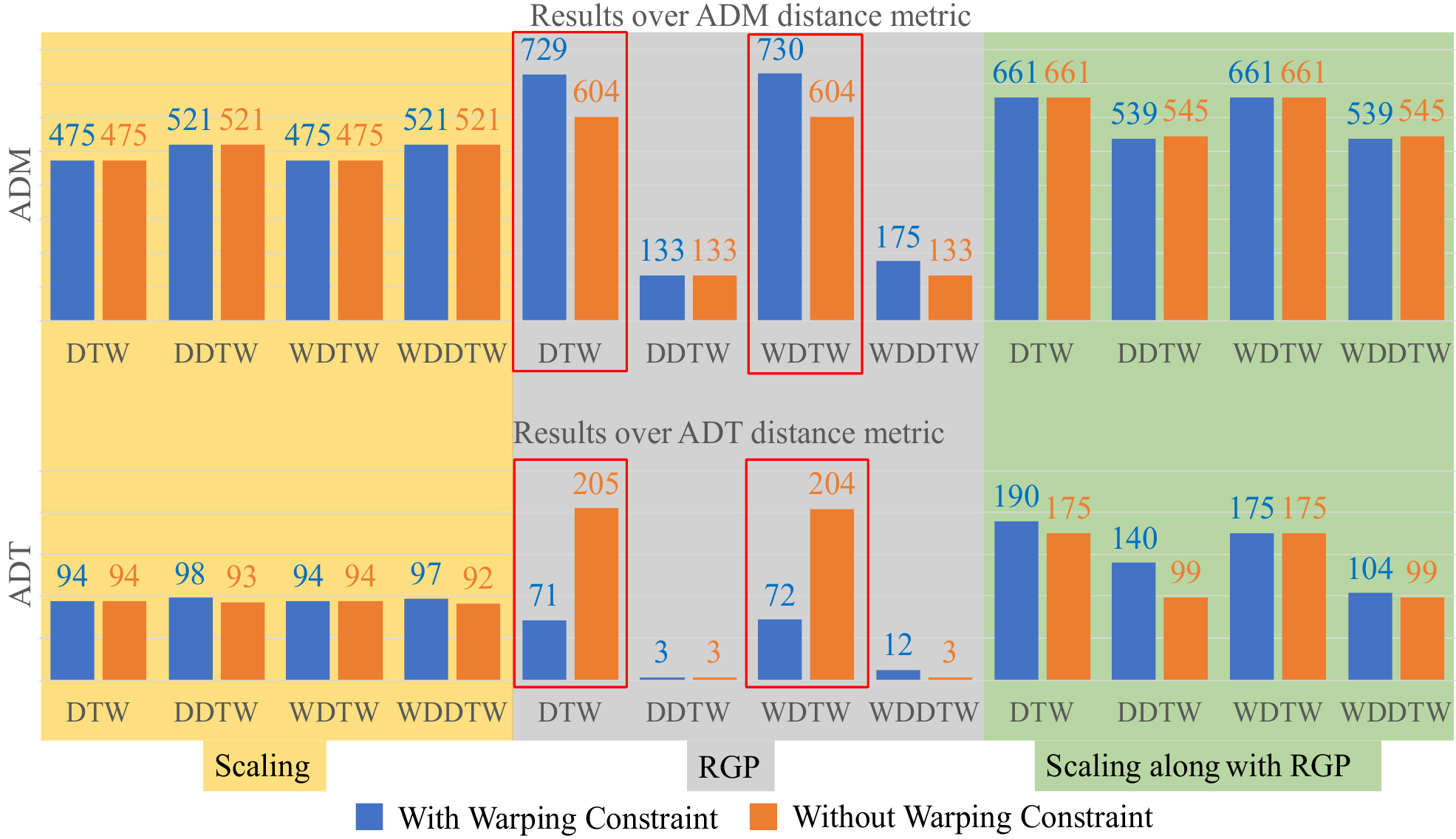}
    \caption{Performance of DTW measures with window constraints. The average distance in 40 signal pairs for each DTW measure is shown.}
    \label{fig:Result_windowing}
\end{figure}

\subsection{DTW measures with Windowing Constraints}
\label{sec:windowing}

Windowing is widely used as the warping constraints on DTW measures. The only parameter there is the width of the window, which is very crucial in terms of the accuracy of the alignment. Figure \ref{fig:Result_windowing} shows experimental results for the evaluation of DTW measures with appropriate warping constraints averaged over 40 time-series pairs. We used synthetic data in these experiments, and we know the true alignment of the reference with the target in each example, hence we set the appropriate width of window for each pair. With window size equal to or slightly larger than scaling, the error rate is almost similar to that of the results \textit{without any warping constraint}. When the time-series pairs have no scaling but random Gaussian peaks (i.e., noise), having a small width of warping window improves the results over the ADT distance metric. Intuitively, because without the warping window constraint the random noise peaks may be aligned with some peaks farther away on the other signal, resulting in low ADM error but high ADT error. This pathological warping can be avoided with the help of window constraints. Detailed visualization of the alignment results can be found in the supplemental document.

\begin{figure*}[th]
    \centering
    \includegraphics[width=0.9\linewidth]{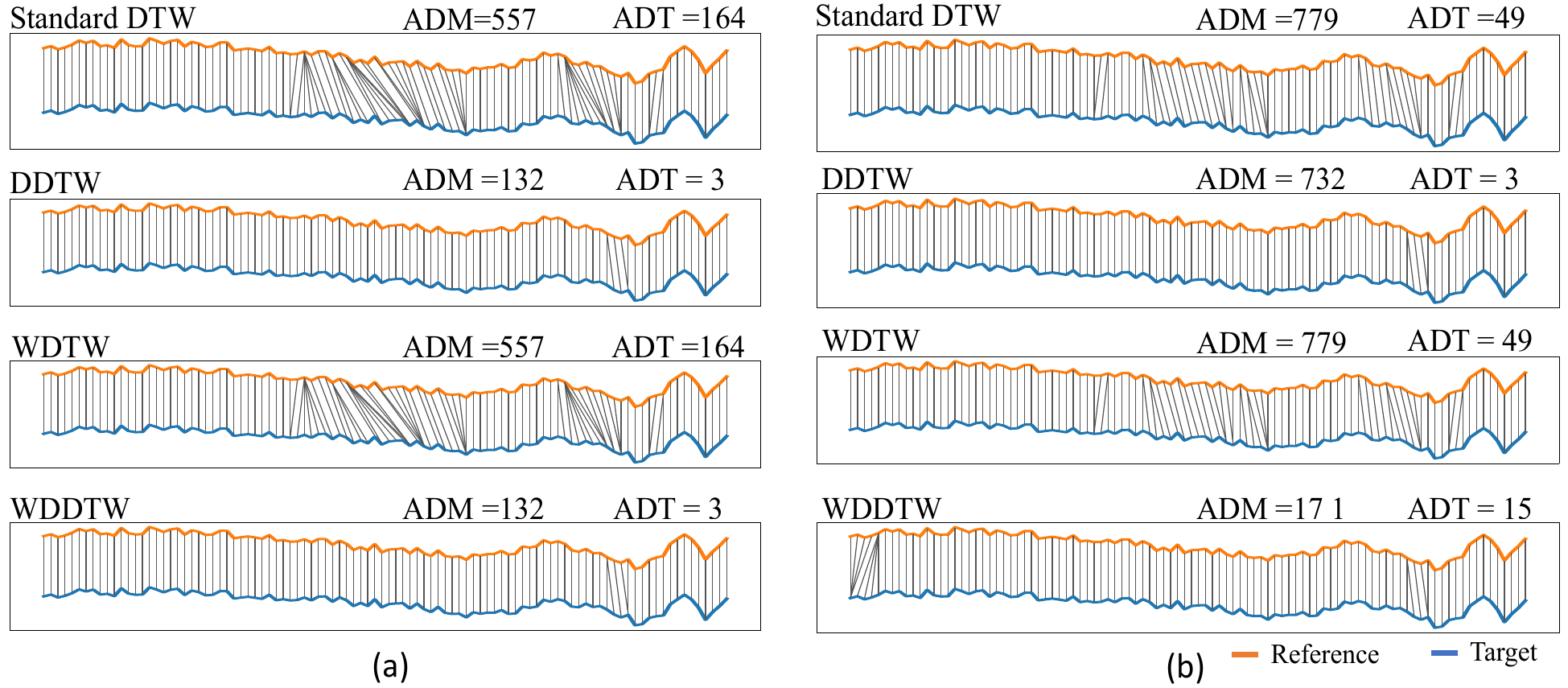}
    \caption{
    Alignment results of DTW measures with (b) and without (a) windowing constraints for the signal pair having variation of random peaks. As we know there is no/very small shifting of the features, we allow a small warping window, it greatly improves the alignment in terms of ADT for DTW and WDTW as shown in (b).
    }
    \label{fig:Windowing_alignment}
\end{figure*}

\begin{figure*}[th]
    \centering
    \includegraphics[width=0.9\linewidth]{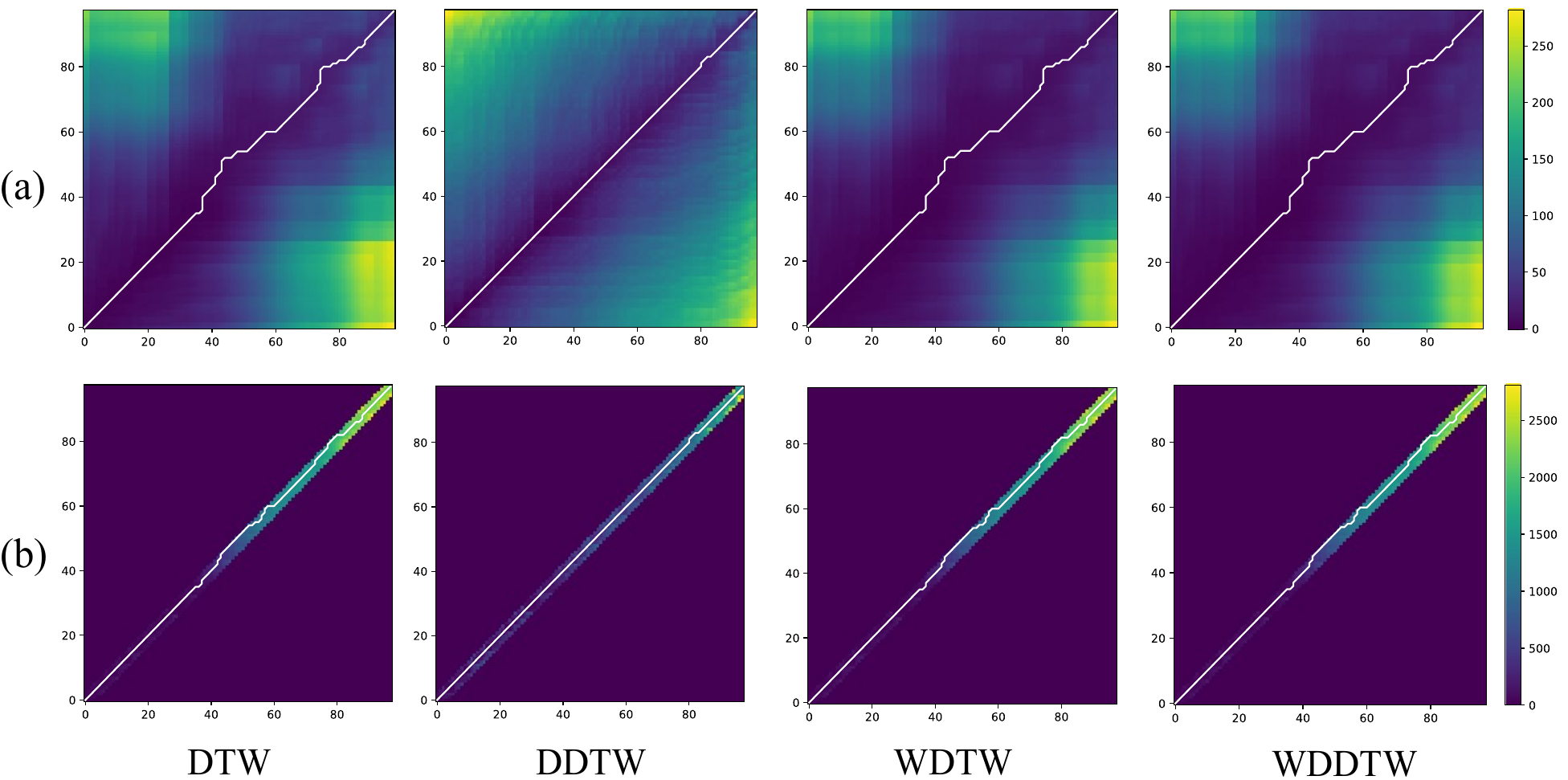}
    \caption{
    Visualization of the warping path for the signal pair instance is shown in Figure \ref{fig:Windowing_alignment}. (a) without warping constraints and (b) with warping constraints. Clearly for DTW and WDTW warping constraints restrict the path from being too skewed.
    }
    \label{fig:Windowing_path}
\end{figure*}




\begin{figure*}[!h]
    \centering
    \includegraphics[width=.95\linewidth]{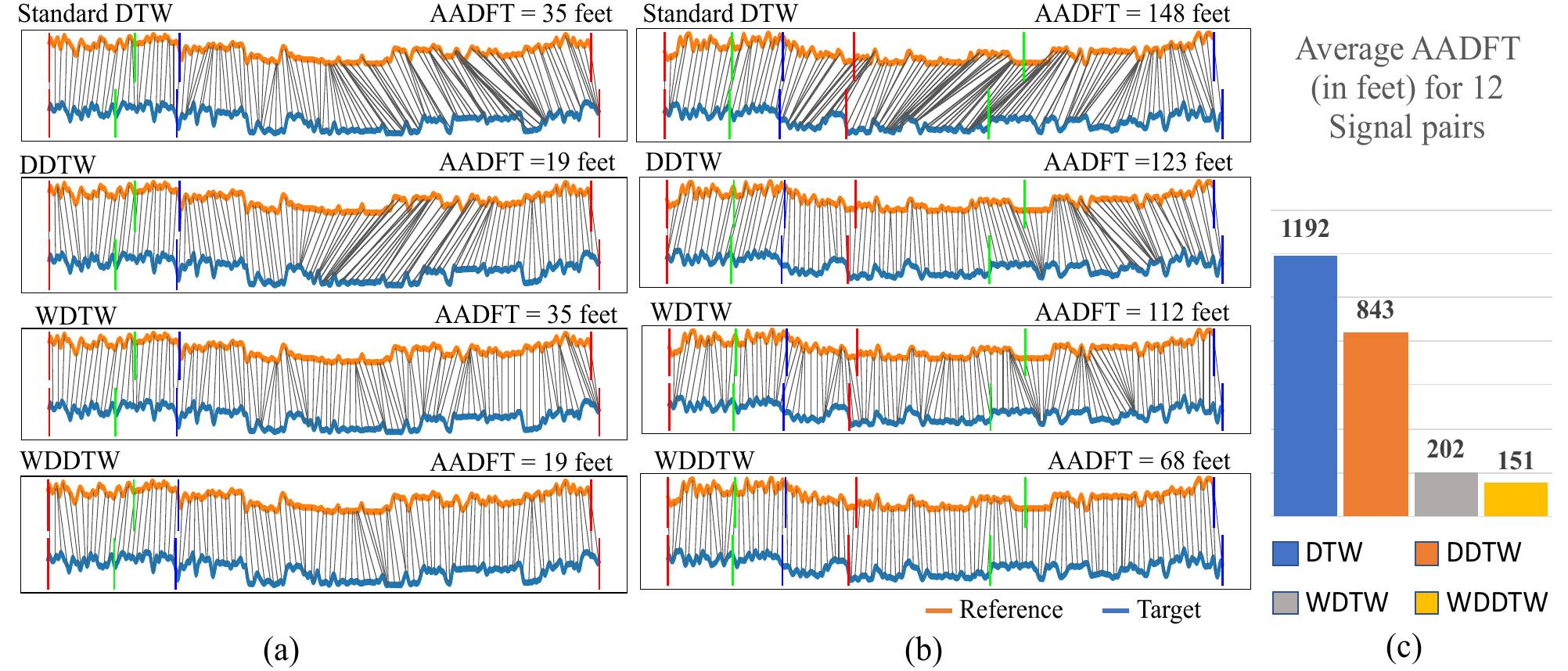}
    \caption{Alignment of Formation Transitions using gamma rays with different DTW measures. (a) and (b) show two different pairs of gamma-ray logs along with their formation transition marked with vertical bars (red, green and blue). The optimized weight parameters are g=0.02 in (a), and g=0.03 in (b) for both WDTW and WDDTW.
    (c) Average AADFT error for different DTW variants over 12 gamma-ray signal pairs, each containing 6 formation transitions.
    }
    \label{fig:Result_SFT}
\end{figure*} 

\subsection{DTWs for Signal Classification}
\label{sec:DTWonClassification}

DTW measures can be used to classify time-series data. The most widely used method in the literature is 1-NN (1 Nearest Neighbor) on \textit{DTW distance} for classification. We generated synthetic data sets to evaluate the performance of different DTW measures on classification.

\subsubsection{Data Set Generation}
\label{subsec:classification_dataset}
To create a data set consisting of $n$ classes, we first generate $n$ linear time-series sequences using our synthesis framework. We then generate different offspring considering these $n$ initial time-series sequences as parents. Each offspring has a random proportion of the signal from all its parents concatenated together and we then perform a few randomly selected deformation operations as described in Figure \ref{sec:ourframework} to further increase the variation. The class label of an offspring is the same as the parent who contributes the most in the generation of this offspring.

\subsubsection{Evaluation Results}
\label{subsec:classification_results}
Based on the experiments performed on the previously described synthetic data set, it is observed that Standard DTW is the winner with 86\% accuracy in labeling a signal to the correct class. WDTW and WDDTW both achieved 73\%, and DDTW achieved 55\% accuracy, respectively. More importantly, we do not observe obvious differences in the performance of different DTW measures on the classification of synthetic time-series data generated with different variations as described in Figure \ref{sec:ourframework}. The reason for this variation in ranking of DTW measures between alignment and classification is that in alignment, local features are given importance while in classification overall (DTW) distance between two signals is considered which is a global attribute over the entire signal. In summary, we believe that \textbf{for signal classification without known characteristics, standard DTW is sufficient}. Our observation is different from what was reported in 
\cite{jeong2011weighted}, which asks for a more in-depth investigation to understand the unique characteristics of the time-series data used in that work.

%% file: Content/Applications.tex
\section{Applications}
\label{sec:applications}


\subsection{Surface Formation transitions alignment}

When drilling in the oil and gas industries, surface formations under the earth play a very crucial role. Often, drilling engineers are interested in knowing the surface formation transition depths in order to fine-tune the drilling parameters. These formation transitions can be found using Gamma ray logs, Resistivity logs and Spontaneous potential logs. Geologists analyze these logs simultaneously to mark the boundaries of formation tops based on their domain knowledge. These signals are recorded from the wells.
Among these three logs, gamma-ray logs are the easiest to record and found in most of the wells, while resistivity and potential logs are not always available. That said, geologists can mark the
formation transitions based on the gamma-ray logs with/without any other additional information (logs). The process of matching the formation transitions by geologists involves logs from one or more wells whose surface formations are known and logs from some other wells whose formations are not known, geologists then try to match the depth level of formation transition from known logs (also called reference) to the logs whose formations are not known (also called target) based on the trends/features of their respective gamma-ray logs.
DTW and its variants can be used to automate this matching process. It can align reference logs with their formation tops marked with one or more other logs (Targets) that do not have formation information. Then, formation transitions in the target logs can be automatically marked based on the alignment with reference logs and reference formation transitions.


\begin{figure*}[th]
    \centering
    \includegraphics[width=.9\linewidth]{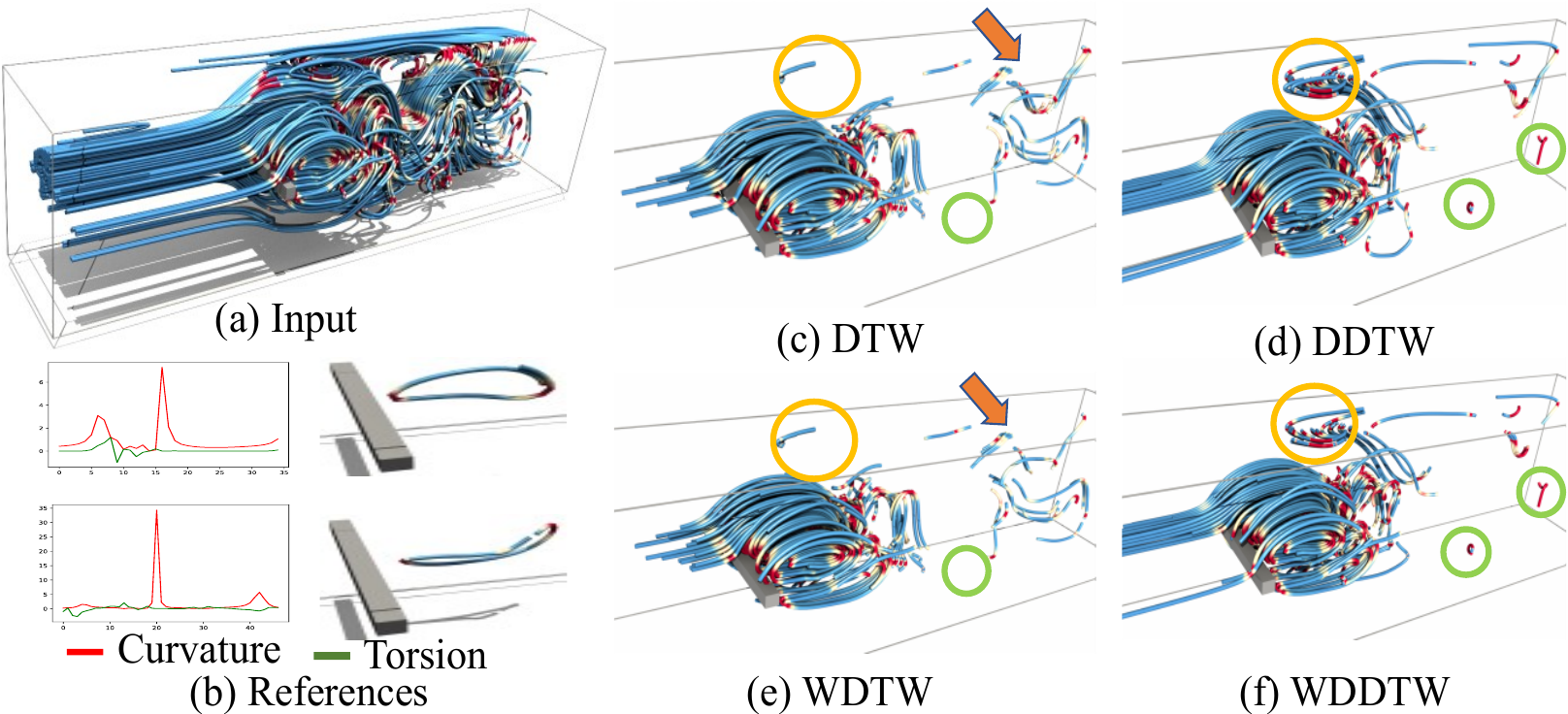}
    \caption{
        Selected streamline segments with characteristics similar to the reference streamlines (b) using different DTW measures (c--f). Colors on streamlines are mapped to the local curvature values. Orange arrows highlight the segments that do not have the desired geometry characteristic as the references (i.e., outliers) and the green circles indicate location with a small-scale feature. Orange circles highlight features that are missed. As can be seen, WDDTW enables the detection of more features similar to the references with fewer outliers.
    }
    \label{fig:Streamlines}
\end{figure*} 

For this type of alignment problems, a better evaluation measure is to find the mismatch in the alignment of important features. Here, we compute the aggregate absolute difference between matched depth and true depth of formation transitions \textbf{(AADFT)}, which is similar to the previously introduced ADT measure. 
Figure \ref{fig:Result_SFT} (a) and (b) show the alignment of two instances of synthetic data which closely resemble gamma-ray logs \footnote{Due to the confidential nature of the data we can not show the real data sequences.}. It is seen that standard DTW and DDTW cannot outperform weighted DTW and weighted DDTW. In some cases, WDTW may outperform DDTW but the difference is not very large.


Figure \ref{fig:Result_SFT} (c) shows the AADFT error value for each of the four DTW variants on 12 such pairs of logs. \textbf{These gamma-ray log pairs have variation which shows both scaling and random peaks. The Weighted Derivative DTW performs the best among others, which is also suggested in our guidelines for this type of variations.} 

\textbf{In order to find the type of difference/variance existing between two signals to-be-aligned}, we can use our Simulated Annealing based parameter fitting framework discussed in Figure \ref{sec:simulatedannealing}. We record the parameters of each operation done by the framework while transforming a signal into the other. The effect of individual scaling as well as addition/subtraction of individual peaks performed by the framework are then evaluated in order to conclude if there exists a significant amount of scaling or peak difference between the two signals. A peak has three parameters: magnitude of the center of the peak, width of the peak, and location of the peak. Both the magnitude and the width of the peak affect the alignment.  The effect of a peak is directly proportional to the product of the width of the peak and the magnitude of the center. It can be mathematically quantified as,
$$Effect \propto (magnitude \times width)$$

We need to normalize this with respect to the amplitude and length of the signal as,
$$Effect = \frac{magnitude \times width}{Amplitude \times length(Signal)} \times 100$$

For multiple iterations of the addition/subtraction of the peaks, the total effect is the sum of the effects created by individual peaks. If the total effect is larger or equal to 5\%, 
then one of the two signals has more non-trivial peaks that do not exist in the other signal.

Similarly, the effect of scaling can be computed by simply considering the amount of change in the length of the scaling window. We can normalize this change in the width of the scaling window with the length of the entire signal. If the sum of the absolute change in length is larger than 5\%, then one of the signals has non-trivial features scaled up/down, as compared to the other. 

The gamma-ray log pairs shown in Figure \ref{fig:Result_SFT} have 12\% and 18\% differences as scaling effect and 9\% and 16\% differences as peaks effect, respectively. The average amount of scaling and peak effects in all the possible pairs in our data set is 11\% and 19\%, respectively. This indicates that the signal pairs in this application possess the combined variations of scaling/shifting and peak difference. Based on our guideline obtained from the evaluation (Section \ref{sec:evl_peaksonscaling}), WDDTW performs the best for this type of variation among signal pairs for alignment. This is confirmed by our assessment reported in Figure \ref{fig:Result_SFT} (c).

\subsection{Pattern Search in Streamlines for Flow Visualization}


Fluid flow can be visualized with the help of streamlines. Though streamlines are 3-dimensional data, their behavior can be described by their curvature and torsion profiles, which are 1-dimensional temporal sequences. Streamlines may represent many interesting features, for example swirling motion in the flow corresponding to vortices. DTW measures can be used to search such features by considering one of the most representative features as a reference. 

Since the interesting patterns may be a portion of a streamline, we perform sliding window DTW on curvature and torsion profiles of the streamlines with the references, while allowing the size of the window to vary between half and double the size of the reference. We consider both the curvature and torsion profiles during the search and compute the Euclidean distance of the two profiles between the reference and target, denoted by $d_c$ and $d_t$. We then combine the two distances using $\sqrt{d_c^2+d_t^2}$. 
The segments of the streamlines with DTW distance smaller than certain thresholds are highlighted in Figure \ref{fig:Streamlines}. We selected a threshold for each DTW measure that highlights the minimum number of outliers. Based on our experiments shown in Figure \ref{fig:Streamlines}, derivative measures (DDTW and WDDTW) work better than others. Specially, WDDTW works the best for our data set. Since the curvature and torsion profile segments vary from references in terms of both scaling and random peaks, it further validates our guidelines that WDDTW works the best for such data. However, none of the DTW measures can capture all features that describe the swirling flow motion. For instance, a bundle of swirling streamline segments near the top boundary (highlighted by the orange circles) are only partially captured by DDTW and WDDTW, while completely missed by DTW and DDTW. This is mainly due to the curvature and torsion configuration at those segments, making their distance to the reference larger than the error threshold (e.g., around 65 for the current results). More representative local attributes or a global descriptor may be needed to fully address this issue, which is beyond the scope of this work.

%% file: Content/Conclusion.tex
\section{Conclusion}
\label{sec:conclusion}


In this work, we introduced a synthesis framework to generate signal pairs with known variations of features. The variation between a generated signal pair is user-controllable. The framework generates a reference signal first and then deforms it to generate the target signal. With this synthesis framework, the correct matching between a pair of signals is known. We then performed extensive experiments for feature alignment on signal pairs with known variations using different variants of Dynamic Time Warping (DTW) methods to assess their performance. This extensive evaluation leads to a first guideline for the selection of the proper DTM measure for the alignment tasks given the characteristics of the signals to be aligned. We verified our guideline with data sets from the applications of surface formation transitions and streamline pattern search. 



\paragraph{Limitations and future works. }
The synthesis framework presented in this work proved to be able to transform a linear signal into another, and it can also fit the parameters to find the existing variations between those two, but it takes many iterations and modifications. We suspect this is due to the use of only Gaussian peaks, and not considering other types of peaks. Using different types of peaks along with Gaussian peaks may improve the efficiency of this framework. Also, our evaluation of alignment of different variations of the signals can be extended to other application-based evaluations with known variations. It can also be extended to other DTW variants like soft-DTW and multi-dimensional DTW. The evaluation can also be extended to the classification and clustering of time-series data using different DTW measures. We plan to explore these directions in the future.

%% file: main.bbl
\begin{thebibliography}{32}
\providecommand{\natexlab}[1]{#1}
\providecommand{\url}[1]{\texttt{#1}}
\expandafter\ifx\csname urlstyle\endcsname\relax
  \providecommand{\doi}[1]{doi: #1}\else
  \providecommand{\doi}{doi: \begingroup \urlstyle{rm}\Url}\fi

\bibitem[Chen et~al.(2001)Chen, Wei, and Zhang]{chen2001discovering}
Guoqing Chen, Qiang Wei, and Hong Zhang.
\newblock Discovering similar time-series patterns with fuzzy clustering and dtw methods.
\newblock In \emph{Proceedings Joint 9th IFSA World Congress and 20th NAFIPS International Conference (Cat. No. 01TH8569)}, volume~4, pages 2160--2164. IEEE, 2001.

\bibitem[Shanker and Rajagopalan(2007)]{shanker2007off}
A~Piyush Shanker and AN~Rajagopalan.
\newblock Off-line signature verification using dtw.
\newblock \emph{Pattern recognition letters}, 28\penalty0 (12):\penalty0 1407--1414, 2007.

\bibitem[Keogh and Pazzani(2001)]{keogh2001derivative}
Eamonn~J Keogh and Michael~J Pazzani.
\newblock Derivative dynamic time warping.
\newblock In \emph{Proceedings of the 2001 SIAM international conference on data mining}, pages 1--11. SIAM, 2001.

\bibitem[Jeong et~al.(2011)Jeong, Jeong, and Omitaomu]{jeong2011weighted}
Young-Seon Jeong, Myong~K Jeong, and Olufemi~A Omitaomu.
\newblock Weighted dynamic time warping for time series classification.
\newblock \emph{Pattern recognition}, 44\penalty0 (9):\penalty0 2231--2240, 2011.

\bibitem[Górecki and Łuczak(2015)]{GORECKI20152305}
Tomasz Górecki and Maciej Łuczak.
\newblock Multivariate time series classification with parametric derivative dynamic time warping.
\newblock \emph{Expert Systems with Applications}, 42\penalty0 (5):\penalty0 2305 -- 2312, 2015.
\newblock ISSN 0957-4174.

\bibitem[Salvador and Chan(2007)]{salvador2007toward}
Stan Salvador and Philip Chan.
\newblock Toward accurate dynamic time warping in linear time and space.
\newblock \emph{Intelligent Data Analysis}, 11\penalty0 (5):\penalty0 561--580, 2007.

\bibitem[Berndt and Clifford(1994)]{berndt1994using}
Donald~J Berndt and James Clifford.
\newblock Using dynamic time warping to find patterns in time series.
\newblock In \emph{KDD workshop}, volume~10, pages 359--370. Seattle, WA, 1994.

\bibitem[Cai et~al.(2019)Cai, Xu, Yi, Huang, and Rajasekaran]{cai2019dtwnet}
Xingyu Cai, Tingyang Xu, Jinfeng Yi, Junzhou Huang, and Sanguthevar Rajasekaran.
\newblock Dtwnet: a dynamic time warping network.
\newblock In \emph{Advances in Neural Information Processing Systems}, pages 11636--11646, 2019.

\bibitem[Cuturi and Blondel(2017)]{cuturi2017soft}
Marco Cuturi and Mathieu Blondel.
\newblock Soft-dtw: a differentiable loss function for time-series.
\newblock In \emph{Proceedings of the 34th International Conference on Machine Learning-Volume 70}, pages 894--903. JMLR. org, 2017.

\bibitem[Izakian et~al.(2015)Izakian, Pedrycz, and Jamal]{izakian2015fuzzy}
Hesam Izakian, Witold Pedrycz, and Iqbal Jamal.
\newblock Fuzzy clustering of time series data using dynamic time warping distance.
\newblock \emph{Engineering Applications of Artificial Intelligence}, 39:\penalty0 235--244, 2015.

\bibitem[Petitjean et~al.(2011)Petitjean, Ketterlin, and Gan{\c{c}}arski]{petitjean2011global}
Fran{\c{c}}ois Petitjean, Alain Ketterlin, and Pierre Gan{\c{c}}arski.
\newblock A global averaging method for dynamic time warping, with applications to clustering.
\newblock \emph{Pattern Recognition}, 44\penalty0 (3):\penalty0 678--693, 2011.

\bibitem[Oates et~al.(1999)Oates, Firoiu, and Cohen]{oates1999clustering}
Tim Oates, Laura Firoiu, and Paul~R Cohen.
\newblock Clustering time series with hidden markov models and dynamic time warping.
\newblock In \emph{Proceedings of the IJCAI-99 workshop on neural, symbolic and reinforcement learning methods for sequence learning}, pages 17--21. Citeseer, 1999.

\bibitem[Jeong and Jayaraman(2015)]{jeong2015support}
Young-Seon Jeong and Raja Jayaraman.
\newblock Support vector-based algorithms with weighted dynamic time warping kernel function for time series classification.
\newblock \emph{Knowledge-based systems}, 75:\penalty0 184--191, 2015.

\bibitem[Mueen(2014)]{mueen2014time}
Abdullah Mueen.
\newblock Time series motif discovery: dimensions and applications.
\newblock \emph{Wiley Interdisciplinary Reviews: Data Mining and Knowledge Discovery}, 4\penalty0 (2):\penalty0 152--159, 2014.

\bibitem[Sakoe and Chiba(1978)]{sakoe1978dynamic}
Hiroaki Sakoe and Seibi Chiba.
\newblock Dynamic programming algorithm optimization for spoken word recognition.
\newblock \emph{IEEE transactions on acoustics, speech, and signal processing}, 26\penalty0 (1):\penalty0 43--49, 1978.

\bibitem[Godin and Lockwood(1989)]{godin1989dtw}
C~Godin and P~Lockwood.
\newblock Dtw schemes for continuous speech recognition: a unified view.
\newblock \emph{Computer Speech \& Language}, 3\penalty0 (2):\penalty0 169--198, 1989.

\bibitem[Bahlmann et~al.(2002)Bahlmann, Haasdonk, and Burkhardt]{bahlmann2002online}
Claus Bahlmann, Bernard Haasdonk, and Hans Burkhardt.
\newblock Online handwriting recognition with support vector machines-a kernel approach.
\newblock In \emph{Proceedings Eighth International Workshop on Frontiers in Handwriting Recognition}, pages 49--54. IEEE, 2002.

\bibitem[Campbell et~al.(1996)Campbell, Becker, Azarbayejani, Bobick, and Pentland]{campbell1996invariant}
Lee~W Campbell, David~A Becker, Ali Azarbayejani, Aaron~F Bobick, and Alex Pentland.
\newblock Invariant features for 3-d gesture recognition.
\newblock In \emph{Proceedings of the second international conference on automatic face and gesture recognition}, pages 157--162. IEEE, 1996.

\bibitem[Faundez-Zanuy(2007)]{faundez2007line}
Marcos Faundez-Zanuy.
\newblock On-line signature recognition based on vq-dtw.
\newblock \emph{Pattern Recognition}, 40\penalty0 (3):\penalty0 981--992, 2007.

\bibitem[Huang and Kinsner(2002)]{huang2002ecg}
Bin Huang and W~Kinsner.
\newblock Ecg frame classification using dynamic time warping.
\newblock In \emph{IEEE CCECE2002. Canadian Conference on Electrical and Computer Engineering. Conference Proceedings (Cat. No. 02CH37373)}, volume~2, pages 1105--1110. IEEE, 2002.

\bibitem[Rath and Manmatha(2003)]{rath2003word}
Toni~M Rath and Raghavan Manmatha.
\newblock Word image matching using dynamic time warping.
\newblock In \emph{2003 IEEE Computer Society Conference on Computer Vision and Pattern Recognition, 2003. Proceedings.}, volume~2, pages II--II. IEEE, 2003.

\bibitem[Gullo et~al.(2009)Gullo, Ponti, Tagarelli, and Greco]{gullo2009time}
Francesco Gullo, Giovanni Ponti, Andrea Tagarelli, and Sergio Greco.
\newblock A time series representation model for accurate and fast similarity detection.
\newblock \emph{Pattern Recognition}, 42\penalty0 (11):\penalty0 2998--3014, 2009.

\bibitem[Fu(2011)]{fu2011review}
Tak-chung Fu.
\newblock A review on time series data mining.
\newblock \emph{Engineering Applications of Artificial Intelligence}, 24\penalty0 (1):\penalty0 164--181, 2011.

\bibitem[Kulbacki and Bak(2002)]{kulbacki2002unsupervised}
Marek Kulbacki and Artur Bak.
\newblock Unsupervised learning motion models using dynamic time warping.
\newblock In \emph{Intelligent Information Systems 2002}, pages 217--226. Springer, 2002.

\bibitem[Zhang et~al.(2014)Zhang, Sun, and Luo]{zhang2014one}
Xianglilan Zhang, Jiping Sun, and Zhigang Luo.
\newblock One-against-all weighted dynamic time warping for language-independent and speaker-dependent speech recognition in adverse conditions.
\newblock \emph{PloS one}, 9\penalty0 (2), 2014.

\bibitem[Wegner~Maus et~al.(2019)Wegner~Maus, C{\^a}mara, Appel, and Pebesma]{wegner2019dtwsat}
Victor Wegner~Maus, Gilberto C{\^a}mara, Marius Appel, and Edzer Pebesma.
\newblock dtwsat: Time-weighted dynamic time warping for satellite image time series analysis in r.
\newblock \emph{Journal of Statistical Software}, 88\penalty0 (5):\penalty0 1--31, 2019.

\bibitem[Maus et~al.(2016)Maus, C{\^a}mara, Cartaxo, Sanchez, Ramos, and De~Queiroz]{maus2016time}
Victor Maus, Gilberto C{\^a}mara, Ricardo Cartaxo, Alber Sanchez, Fernando~M Ramos, and Gilberto~R De~Queiroz.
\newblock A time-weighted dynamic time warping method for land-use and land-cover mapping.
\newblock \emph{IEEE Journal of Selected Topics in Applied Earth Observations and Remote Sensing}, 9\penalty0 (8):\penalty0 3729--3739, 2016.

\bibitem[Thawonmas and Iizuka(2008)]{thawonmas2008visualization}
Ruck Thawonmas and Keita Iizuka.
\newblock Visualization of online-game players based on their action behaviors.
\newblock \emph{International Journal of Computer Games Technology}, 2008, 2008.

\bibitem[Zhu et~al.(2012)Zhu, Kim, Proctor, Narayanan, and Nayak]{zhu2012dynamic}
Yinghua Zhu, Yoon-Chul Kim, Michael~I Proctor, Shrikanth~S Narayanan, and Krishna~S Nayak.
\newblock Dynamic 3-d visualization of vocal tract shaping during speech.
\newblock \emph{IEEE transactions on medical imaging}, 32\penalty0 (5):\penalty0 838--848, 2012.

\bibitem[Hachaj et~al.(2017)Hachaj, Ogiela, Piekarczyk, and Koptyra]{hachaj2017advanced}
Tomasz Hachaj, Marek~R Ogiela, Marcin Piekarczyk, and Katarzyna Koptyra.
\newblock Advanced human motion analysis and visualization: comparison of mawashi-geri kick of two elite karate athletes.
\newblock In \emph{2017 IEEE Symposium Series on Computational Intelligence (SSCI)}, pages 1--7. IEEE, 2017.

\bibitem[Lee and Shen(2009)]{lee2009visualization}
Teng-Yok Lee and Han-Wei Shen.
\newblock Visualization and exploration of temporal trend relationships in multivariate time-varying data.
\newblock \emph{IEEE Transactions on Visualization and Computer Graphics}, 15\penalty0 (6):\penalty0 1359--1366, 2009.

\bibitem[Van~Laarhoven and Aarts(1987)]{van1987simulated}
Peter~JM Van~Laarhoven and Emile~HL Aarts.
\newblock Simulated annealing.
\newblock In \emph{Simulated annealing: Theory and applications}, pages 7--15. Springer, 1987.

\end{thebibliography}
